# Indian hieroglyphs

## ■ Indus script corpora, archaeo-metallurgy and Meluhha (Mleccha)

Jules Bloch's work on formation of the Marathi language (Bloch, Jules. 2008, Formation of the Marathi Language. (Reprint, Translation from French), New Delhi, Motilal Banarsidass. ISBN: 978-8120823228) has to be expanded further to provide for a study of evolution and formation of Indian languages in the Indian language union (*sprachbund*). The paper analyses the stages in the evolution of early writing systems which began with the evolution of counting in the ancient Near East. Providing an example from the Indian Hieroglyphs used in Indus Script as a writing system, a stage anterior to the stage of syllabic representation of sounds of a language, is identified. Unique geometric shapes required for tokens to categorize objects became too large to handle to abstract hundreds of categories of goods and metallurgical processes during the production of bronze-age goods. In such a situation, it became necessary to use glyphs which could distinctly identify, orthographically, specific descriptions of or cataloging of ores, alloys, and metallurgical processes. About 3500 BCE, Indus script as a writing system was developed to use hieroglyphs to represent the 'spoken words' identifying each of the goods and processes. A rebus method of representing similar sounding words of the lingua franca of the artisans was used in Indus script. This method is recognized and consistently applied for the lingua franca of the Indian *sprachbund*. That the ancient languages of India, constituted a *sprachbund* (or language union) is now recognized by many linguists. The *sprachbund* area is proximate to the area where most of the Indus script inscriptions were discovered, as documented in the corpora. That hundreds of Indian hieroglyphs continued to be used in metallurgy is evidenced by their use on early punch-marked coins. This explains the combined use of syllabic scripts such as Brahmi and Kharoshti together with the hieroglyphs on Rampurva copper bolt, and Sohgaura copper plate from about 6[th] century BCE. Indian hieroglyphs constitute a writing system for meluhha language and are rebus representations of archaeo-metallurgy lexemes. The rebus principle was employed by the early scripts and can legitimately be used to decipher the Indus script, after secure pictorial identification.

Invention of bronze-age technologies necessitated the invention and development of a writing system called Indus Script which is evidenced in corpora of about 6000 inscriptions.[1] Around 7500 BCE[2], tokens appeared and represented perhaps the early deployment of a writing system to count objects. Many geometric shapes were used for the tokens.[3] Tracing the evolution of a writing system[4], Schmandt-Besserat evalutes the next stage of keeping tokens in envelopes with markings abstracting the tokens inside and calls these abstract numbers are 'the culmination of the process…'[5] This evaluation is the starting point for identifying another stage before 'the culmination' represented by the use of syllabic representation in glyphs of sounds of a language.

That penultimate stage, before syllabic writing evolved, was the use of hieroglyphs represented on hundreds of Indian hieroglyphs.[6]

The arrival of the bronze age was maked by the invention of alloying copper with arsenic, zinc or tin to produce arsenic-alloys, and other alloys such as brass, bronze, pewter. These archaeo-metallurgial inventions enabled the production of goods surplus to the requirements of the artisan guilds. These inventions also created the imperative of and necessity for a writing system which could represent about over 500 specific categories of activities related to the artisanal repertoire of a smith. Such a large number of categories could not be handled by the limited number of geometric shapes used in the token system of accounting and documenting – goods, standard measures of grains, liquids and surface areas.[7]

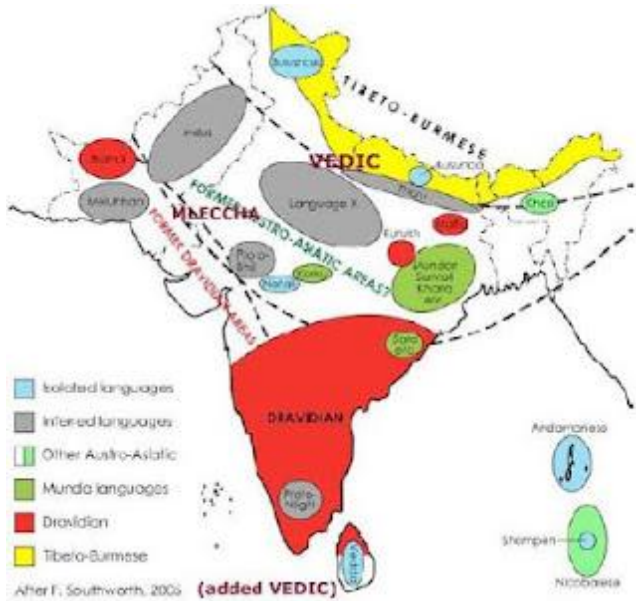

The existence of Indian *sprachbund* is evidenced by the concordant lexemes used for bronze-age repertoire of bronze-age artisans. These lexemes are compiled in an *Indian Lexicon*.[8] This is a resource base for further studies in the formation and evolution of most of the Indian languages. Identifiable substrata glosses include over 4000 etyma of Dravidian Etymological Dictionary and over 1000 words of Munda with concordant semantic clusters of Indo-Aryan. That the substrata glosses cover three major language families –Dravidian, Munda and Indo-Aryan -- is a surprising discovery. There are over 1240 semantic clusters included in the *Indian Lexicon* from over 25 languages which makes the work very large, including cognate entries of CDIAL (Indo-Aryan etyma), together with thousands of lexemes of Santali, Mundarica and other languages of the Austro-Asiatic linguistic group, and, maybe, Language X. . Most of the lexical archive relate to the bronze-age cultural context and possible entries are relatable rebus to Indian hieroglyphs. Many are found to be attested as substratum lexemes only in a few languages such as Nahali, Kashmiri, Kannada or Telugu or lexical entries of Hemacandra's *deśī nāmamālā (Prākṛt)*; thus, many present-day Indian languages are rendered as dialects of an Indus language or proto-Indic *lingua franca* or gloss. The identification of a particular Indian language as the Indus language has presented some problems because of the received wisdom about grouping of language families in Indo-European linguistic analyses. Some claims of decipherment have assumed the language to be Tamil, of Dravidian language family; some have assumed the language to be Sanskrit, of Indo-Aryan language family. A resolution to these problems comes from a surprising source: Manu.

Mleccha, Indus language of Indian linguistic area (*sprachbund*). Indian linguistic area map, including mleccha and vedic (After F. Southworth, 2005; VEDIC AND MLECCHA added.) A language family, mleccha (?language X), is attested in the ancient literature of India. This is the lingua franca, the spoken version of the language of the civilization of about 5000 years ago, distinct from the grammatically correct version called Sanskrit represented in the vedic texts and other ancient literature. Ancient texts of India are replete with insights into formation and evolution of languages. Some examples are: Bharata's Natya Shastra, Patanjali's Mahabhashya, Hemacandra's Deśīnāmamālā, Nighaṇṭus, Panini's Aṣṭādhyayi, Tolkappiyam–Tamil grammar. The evidence which comes from Manu, dated to ca. 500 BCE. Manu (10.45) underscores the linguistic area: ārya vācas mleccha vācas te sarve dasyuvah smṛtāh [trans. "both ārya speakers and mleccha speakers (that is, both speakers of literary dialect and colloquial or vernacular dialect) are all remembered as dasyu"]. Dasyu is a general reference to people. Dasyu is cognate with dasa, which in Khotanese language means 'man'. It is also cognate with daha, a word which occurs in Persepolis inscription of Xerxes, a possible reference to people of Dahistan, a region east of Caspian sea. Strabo wrote :"Most of the scythians, beginning from the Caspian sea, are called Dahae Scythae, and those situated more towards the east Massagetae and Sacae." (Strabo, 11.8.1). Close to Caspian Sea is the site of Altyn-tepe which was an interaction area with Meluhha and where three Indus seals with inscriptions were found, including a silver seal showing a composite animal which can be called a signature glyph of Indus writing..

The identification of mleccha as the language of the Indus script writing system is consistent with the following theses which postulate an Indian linguistic area, that is an area of ancient times when various language-speakers interacted and absorbed language features from one another and made them their own: Emeneau, 1956; Kuiper, 1948; Masica, 1971; Przyludski, 1929; Southworth, 2005.

**Semantic clusters in Indian Lexicon** (1242 English words and Botanical species Latin)

**Economic Court: Flora and Products from Flora**

Birds

Insects

Fauna

Animate phenomena: birth, body, sensory perceptions and actions

Visual phenomen, forms and shapes

Numeration and Mensuration

Economic Court: Natural phenomena, Earth formations, Products of earth (excluding flora clustered in a distinct category)

Building, infrastructure

Work, skills, products of labour and workers (fire-worker, potter/ smith/ lapidary, weaver, farmer, soldier)

Weapons and tools

Language fields

Kinship

Social formations

**Economic Court: Flora and Products from Flora**

butter  curdle  flesh  flour  food  grain  honey  liquor  mahua  molasses  oil  oilcake   rice  spice  sugar  supper  tobacco  wheat bark  cloth  cotton  drug   flax  fragrance   fringe  garland  harvest  granary  gluehemp  indigo  itch  kunda  lac  log   medicine  mouldy  ointment  peel  poisonpulp  pungent  raw  reed  resin  root  sandal  scent  seed  sheaf  sheath  skein  sow  stick  straw  thorn  thresh  tip-cat apple  asparagus  balsam  bamboo  banana  barley  basil  basket  betel  bud  camphor  cardamom  cashew  celery  chaff  clearingnut  clove bush  corkcoconut  coffee  creeper  cucumber  cumin  ebony  date  fenugreek  forest flower   fruit  garden  garlic  ginger  gooseberry  gourd  hibiscus  jackfruit  jalap  jujube

| |
|---|
| leadwort  leaf  linseed  lotus  mango   mushroom  mustard  palm  orpiment  pepper  pericarp  petal  pomegranate  raspberry  saffron  sago  sprout  tree  tuber   turmeric  wax  wood-apple |
| abies  abrus  acacia  acalypha  acampe  acanthus  achyranthes  aconitum  acorusadenanthera  aegle  aeschynomena  aeschynomene  agaricus  agathotes  agati  ageratum  aglaia  aguilaria  ailantus  alangium  aloe  alosanthes  alpinia  amarantus  albizzia  amomum  andropogon  anethum  anodendron  anogeissus  anthocephalus  anthriscus  antiaris  areca  aristolochia  arka  artemisia  artocarpus  arum  atlantia  averrhoea  azima  balanites  barleria  barringtonia  basella  bassia  bauhinia  berberis  betula  bixa  blyxa  bombax  boswelliabryonia  buchanania   butea  caesalpinia  caesaria  cajanus  calamus  calophyllum  canarium  cannabis  canthium  capparis  carallia  cardiospermum   careya  carissa  carthamus  carum  caryota  cassia  cassytha  cedrela  cedrus  celastrus  celosia  celtis  cerbera  ceropegia  ceratonia  chenopodium  cicer  cichorium  cinnabar  cinnamomum  cinnamon  citrus  clarion  cleistanthus  clerodendrum  clitoria  coccinia  cocculus  colocasia  colosanthes  convolvulus  cordia  coriandrum  costum  costus  cratraeva  crocus  crotalaria  croton  cucumis   curculigo  curcuma  cyperus  dalbergia  datura  desmodium  dichrostachys  dillenia  dioscorea  diospyros  dodonea  dolichos  eclipta  elaeocarpus  elettaria  eleusine  ericybe  erythrina  erythroxylon  eugenia  eugenis  euphorbia  excoecaria  feronia  ferula  ficus  frankincense  flacourtia  garcinia  galangal  gambogge  gardenia  gaultheria  gendarussa  gentiana  gloria  gmelina  grewia  grislea  gymnema  gynandropsis  gyrocarpus  heliotropium  hemidesmus  hiptage  holcus  hopea  hydnocarpus  ichnocarpus  ilex  indigofera  ipomoea  jasminum  juniper  justicia  kaempferia  lagenaria  lagerstroemialaurus  lepidum  leucas  ligusticum  linum  lobellia  lodhra  luffa  luvunga  macaranga  mangifera  marsilia  melastoma  meliosma  memecyclon  mentha   mesua  millingtonia  mimusops  momordica  moringa  morus  mucuna  myrica  myristica  myrobalan  myrtus  nardostachys  nauclea  nelumbium  neriumnyctanthes  nymphaea  ochlandra  ochre  ocimum  odina   olea  ophioxylon  oryza  palmyra  pandanus  panicum  papaver  pavetta  pavonia  phaseolus  phoenixphyllanthus  physalis  pimpinella  pinus  piper  plumbago  pogostemon  polygala  polygonum  premna  prunus  psidium  pterocarpus  pterospermum  pouzolzia  prosopis   quercus  randia  raphanus  rauwofia  rhizophora  ricinus  rottleriarubia  rumex  saccharum  sal  salicornia  salvadora  salvinia  sandoricum  santalum  sapindus  sarcostemma  saussurea  schleichera  scirpus  semecarpus   sesamum  sesbana  sesbania  shorea  sinapis  solanum  soymida  sphaeranthus   spinachia  sterculia  stereospermum  strobilanthes  strychnos  swertiasymplocos  syzygium  tabernaemontana  tamarindus  tectona  tephrosia  terminalia  thespesia  tinospora  tribulus  tragia  trapa  trema  trichosanthes  trigonella  trophis  unquis  utrica  vaccinium  veronia  vitex  vulpes  wrightia  xylia  zizyphus |

**Birds**

| |
|---|
| bird  bluejay  cock  crane  crow  cuckoo  dove   duck  eagle  feather   gizzard  crest  hawk  heron  kingfisher  myna   nest  owl  parrot   pheasant  quail  robin  shrikeskylark  snipe  sparrow  teal  weaver-bird |

**Insects**

| |
|---|
| bat  beehive  caterpillar  chameleon  cockroach  crab   frog  insect  lizard  mosquito   scorpion   snake  spider |

**Fauna**

| |
|---|
| animal  antelope, goat, deer, markhor,    ram  alligator  bear  buffalo  bull  camel  dog  elephant   fish  hare  herd  horn   horse  ivory   lair  lion  lowing  mongoose  monkey   musk- |
| deer  octopus pony  porpoise rat rhinoceros  shoal squirrel tail tiger tortoise yak  yak-tail |

**Animate phenomena: birth, body, sensory perceptions and actions**

| |
|---|
| abortion age amazed anger anus arrive ask attack back bald  bathe behindbeard  beat beg being  belly bile  birth bite  blink  blood  blow body  boil bone  breath bristle  butt buttock care  cheek  chest chignon  chin climb  come copulate  creep cross cry cut dance death  decay doubt  dream dumb  dwarf echo elbow  end excrement eye  faeces fall fat  finger fist flee fly  frolic front  funeralgenus  give gore groan hair hand  hatch head  heel  hear  heart herpes hiccup hide hit hunt  hurt idle  intoxicate invite  itch jaundice joint juggler jump kick lame  laugh  lift  leap leg lip listen  liver  look  male mane meet mole mouth  movement  muscle nail navel neck nerve  noise  nose  numb  old  penis perish phlegm plague pour pregnant pudendum pull  pus  push  putraise  rattle  recite reply repress restrain  rinse roar  roll  run  rush scab scar  scatter  seize senses  separate  serve silence sing sink  sit shouldershrink slander  slap  sleep speak  splash spleen split  sprain  squeeze stammer standing  stay stirring stop strength  suck  surprise  swallow  sweep  swell  swing syllable  take tame  taste throb throw tired  toe  trunk tumour   turning turn-back tusk twist udder urine vault vomit vulva waist walkwoman  word  wrinkle  young |

**Visual phenomena, forms and shapes**

| |
|---|
| ball beauty bend  bit black braid brown  bubble chequered circle colour crack curve dense dot endless entangled extremity fitting flow fork full  green  heap hole  hollow hump  incline invert knob knot leak left line  long  loose middle ooze  red  slack slant small square straight stripe white |

**Numeration and Mensuration**

| |
|---|
| account  agreement audit average  balance (scales) banker  big broad center    cheap coin collect  collection contain counting  deficient divide eight finger five  four half high increase joint  knot lightness load mark  marked  market marking numeration one  remainder  six seven ten two three twelve twenty   measure weight zero |

**Economic Court: Natural phenomena, Earth formations, Products of earth (excluding flora clustered in a distinct category)**

| |
|---|
| barren  basin borax brass bright bronze burst clay cloud cold collyriumcrystal darkness dawn desert  dew dry extinguish  fire frost gem glittergold (including soma) goods earth hail heat hill island  lapislazuli  lightningmoon mud  night  north  ocean ore pearl planet  pleiades rain  rainbow river ruby sand salt sediment shell silk  silver sky  smoke soap  solstice south star stone sun tank tin thunder water wave wet wind zodiac |

**Building, infrastructure**

| |
|---|
| arch brick bridge building bund  cave chisel  chop  churn corner door drain fence fencing   ford fort  house  kitchen lattice loft parapet pillar rafter roof    shelf  space stable wall wattle  way (path, road) |

Work, skills, products of labour and workers (fire-worker, potter/ smith/ lapidary, weaver, farmer, soldier)

[The lexemes related to weapons and tools are so vivid and distinctive that the entire group has been clustered together to provide an overview of the skills developed which are reflected in semantic expansions related to weapon types and to wielding them. Thus, the clusters in the following list (e.g. awl, axe, bow, goad, razor, saw, sickle) are only to be treated as 'tool' samplers of a Metals Age, emerging out of a lithic age.]

**Weapons and Tools**

awl  axe bow goad razor saw sickle assembly amulet army  axle  badge  bead  bed bellows  blanket  boat  bolt  bore  bracelet  brazier  break  broken  butcher  camp  cart  carve  censer  cloak  comb  commonwealth  convey  crucible  cymbals deliver  dent  depart  dice  distill  drill  drive  drum  edge  embark  engrave  enter  entreat  erect
fan  fasten fatigue  fear  fell  ferry  filter  fire  flag  flute  forge  fry furnace  furrow  glove  gong   groove  guard  guild  hammer  indra jacket
join  kill  kiln  kubera  labour  ladder  ladle  lamp  land  landless  lathe  leash  leather  lid  lever  loom  lute manger  mill  mirror  mould  necklace  net occupation  oil-press  ornament  pannier  patrol  perforate  pin  plait  plough  pole  pot  potsherd  potter  pressed  produce  profession  pure  purity  raft  rope  screen  seat  sew  shackle  script  sling (bearing/carrying)  snare  soldier  spike spinner (weaver)  spy  stake  stamp steam  stirrup   stool  stopper  store  tablet   trap  treasury  trough  uproot  vessel  warrior  wash  water-lift  well  well-digger wheel  whip  winnow  write

**Language fields**

| grammar (Etymology, linguistics, grammar, particles, prepositions, adverbs) | arab tamil telugu<br><br>become near next now  only other that there thus time until<br><br>augment consonant name  no prefix riddle sign signature     yes |
|---|---|

**Kinship**

ancestor  bride brother companion family  father friend gentleman girl lead love marriage mistress  mother    self  single  sister  wife

**Social formations**

abuse  ambush  auction  authentic bard  bawd  brahma   bravo  buy  chief  class commend  confidence  conflict    confusion  cruel  country  court  dedicate  deity  demon  disgrace  doctrine  evil  exile  faith  festival  fop  fraud  free  freedom  game  get  gift  goblin  good  gratitude  guilt  hindu   honour  idol  justice  law  learn  lease  lend  life load  loan  malice  manner market  meditate  memorial   mercy  miser  mystic  oppose  painting  penalty  place  play  please  pledge  pomp  poor  post  power  prank  pride  principal  procession  protect  regularity  regulation  rich  rob  rogue  royal  rule  sacrifice  safety  salutation scheme  sell  send  shame  sindhu  stupid  support  surplus  tax  teacher  temple  terror  theft  tomb  town  trade  tribe  unruly  useless  value  violence virtuous  vow  wager  wicked  win  witness  worship

**Other semantic clusters (including cognisance and lexemes which may indicate semantic expansion and may span many other semantic clusters; e.g. 'mix' cluster may relate to animate and inanimate clusters)**

adhere  begin  blocked  bold  bundle  clean  clever  close  coax    commence dangle  deceit  defeat  deliberate  desire  detached  dip  dirty  disgust  dull  enclose  endure  false  forget  hard  inferior  know  mark  marked  marking mass  means  medley  mix   narrow  neat  nee d  new  notch  opportunity  outside  overflow  part  particle  paste  pit  pitfall  ponder  purpose  quick  quit  ready  remember  rise  rot  rough  rub  ruin  section  shade  shake  similar slow  strip  thin  think  trace  tranquil  trouble  truth  unripe  upper  vermillion

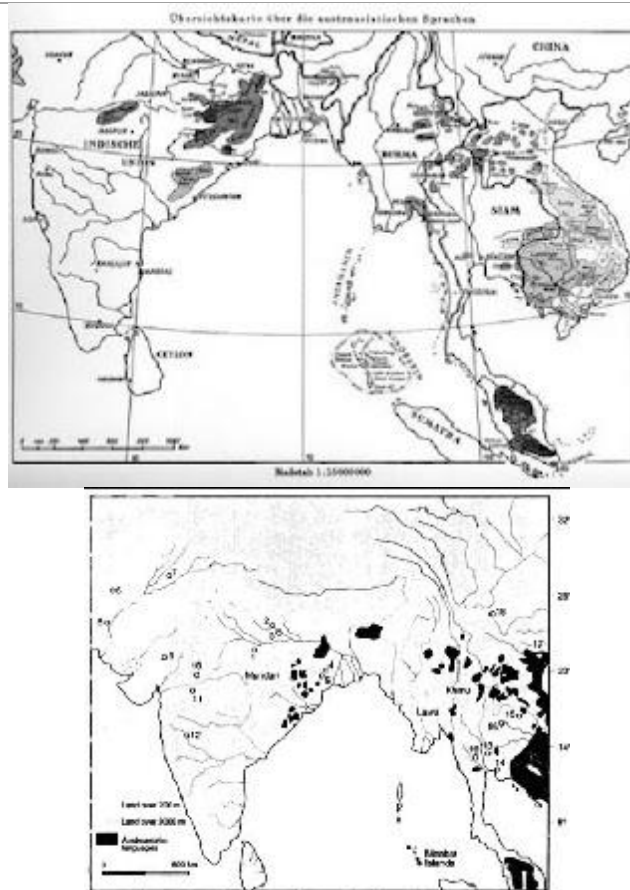

Pinnow's map of Austro-AsiaticLanguage speakers correlates with bronze age sites. http://www.ling.hawaii.edu/faculty/stampe/aa.html   See http://kalyan97.googlepages.com/mleccha1.pdf  The areal map of Austric (Austro-Asiatic languages) showing regions marked by Pinnow correlates with the bronze age settlements in Bharatam or what came to be known during the British colonial regime as 'Greater India'. The bronze age sites extend from Mehrgarh-Harappa (Meluhha) on the west to Kayatha-Navdatoli (Nahali) close to River Narmada to Koldihwa- Khairdih-Chirand on Ganga river basin to Mahisadal – Pandu Rajar Dhibi in Jharia mines close to Mundari area and into the east extending into Burma, Indonesia, Malaysia, Laos, Cambodia, Vietnam, Nicobar islands. A settlement of Inamgaon is shown on the banks of River Godavari.

Bronze Age sites of eastern India and neighbouring areas: 1. Koldihwa; 2.Khairdih; 3. Chirand; 4. Mahisadal; 5. Pandu Rajar Dhibi; 6.Mehrgarh; 7. Harappa;8. Mohenjo-daro; 9.Ahar; 10. Kayatha; 11.Navdatoli; 12.Inamgaon; 13. Non PaWai; 14. Nong Nor;15. Ban Na Di andBan Chiang; 16. NonNok Tha; 17. Thanh Den; 18. Shizhaishan; 19. Ban Don Ta Phet [After Fig. 8.1 in: Charles Higham, 1996, The Bronze Age of Southeast Asia,  Cambridge University Press].

**Evidence related to proto-Indian or proto-Indic or Indus language**

A proto-Indic language is attested in ancient Indian texts. For example, Manusmṛti refers to two languages, both of dasyu (daha): ārya vācas, mleccha vācas. *mukhabāhū rupajjānām yā loke jātayo bahih mlecchavācas'cāryav ācas te sarve dasyuvah smṛtāh* Trans. 'All those people in this world who are excluded from those born from the mouth, the arms, the thighs and the feet (of Brahma) are called Dasyus, whether they speak the language of the mleccha-s or that of the ārya-s.' (Manu 10.45)] This distinction between *lingua franca* and literary version of the language, is elaborated by Patañjali as a reference to 1) grammatically correct literary language and 2) ungrammatical, colloquial speech (*deśī*).

Ancient text of Panini also refers to two languages in *śikṣā*: Sanskrit and Prākṛt. Prof Avinash Sathaye provides a textual reference on the earliest occurrence of the word, 'Sanskrit':

*triṣaṣṭiścatuh ṣaṣṭirvā varṇāh ṣambhumate matāh |*

*prākṛite samskṛte cāpi svayam proktā svayambhuvā || (pāṇini's śikṣā)*

Trans. There are considered to be 63 or 64 varṇā-s in the school (mata) of shambhu. In Prakrit and Sanskrit by swayambhu (manu, Brahma), himself, these varṇā-s were stated.

This demonstrates that pāṇini knew both samskṛta and prākṛita as established languages. (Personal communication, 27 June 2010 with Prof. Shrinivas Tilak.)

Chapter 17 of Bharatamuni's *Nāṭyaśāstra* is a beautiful discourse about Sanskrit and Prakrit and the usage of *lingua franca* by actors/narrators in dramatic performances. Besides, Raja Shekhara, Kalidasa, Shudraka have also used the word Sanskrit for the literary language. (Personal communication from Prof. TP Verma, 7 May 2010). *Nāṭyaśāstra* XVII.29-30: *dvividhā jātibhāṣāca prayoge samudāhṛtā mlecchaśabdopacārā ca bhāratam varṣam aśritā* 'The jātibhāṣā (common language), prescribed for use (on the stage) has various forms. It contains words of mleccha origin and is spoken in Bhāratavarṣa only…' Vātstyāyana refers to mlecchita vikalpa (cipher writing of mleccha) Vātstyāyana's Kamasutra lists (out of 64 arts) three arts related to language:

- *deśa bhāṣā jñānam* (knowledge of dialects)
- *mlecchita vikalpa* (cryptography used by mleccha) [cf. mleccha-mukha 'copper' (Skt.); the suffix –mukha is a reflex of mūh 'ingot' (Mu.)
- *akṣara muṣṭika kathanam* (messaging through wrist-finger gestures)

Thus, semantically, mlecchita vikalpa as a writing system relates to cryptography (perhaps, hieroglyphic writing) and to the work of artisans (smiths). I suggest that this is a reference to Indian hieroglyphs.

It is not a mere coincidence that early writing attested during historical periods was on metal punch-marked coins, copper plates, two-feet long copper bolt used on an Aśokan pillar at Rampurva, Sohoura copper plate, two pure tin-ingots found in a shipwreck in Haifa, and even on the Delhi iron pillar clearly pointing to the smiths as those artisans who had the competence to use a writing system. In reference to Rampurva copper-bolt: "Here then these signs occur upon an object which must have been made by craftsmen working for Asoka or one of his predecessors." (F.R. Allchin, 1959, Upon the contextual significance of certain groups of ancient signs, *Bulletin of the School of Oriental and African Studies*, London.). The Indus script inscriptions using hieroglyphs on two pure tin-ingots found in Haifa were reviewed (Kalyanaraman, S., 2010, The Bronze Age Writing System of Sarasvati Hieroglyphics as Evidenced by Two "Rosetta Stones" - Decoding Indus script as repertoire of the mints/smithy/mine-workers of Meluhha. *Journal of Indo-Judaic Studies*. Number 11. pp. 47–74).

*Mahābhārata* also attests to mleccha used in a conversation with Vidura. *Śatapatha Brāhmaṇa* refers to mleccha as language (with pronunciation variants) and also provides an example of such mleccha pronunciation by asuras.  A Pali text, *Uttarādhyayana Sūtra* 10.16 notes: *ladhdhaṇa vimānusattaṇṇam āriattam puṇrāvi dullaham bahave dasyū milakkhuyā*; trans. 'though one be born as a man, it is rare chance to be an ārya, for many are the dasyu and milakkhu'. Milakkhu and dasyu constitute the majority, they are the many. Dasyu are milakkhu (mleccha speakers). Dasyu are also ārya vācas (Manu 10.45), that is, speakers of Sanskrit. Both ārya vācas and mleccha vācas are dasyu [cognate *dahyu, daŋha, daha* (Khotanese)], people, in general. दाशः 1 A fisherman; इयं च सज्जा नौश्चेति दाशाः प्राञ्ज- लयो$ब्रुवन् Rām.7.46.32; Ms.8.48,49;1.34. दासः 'a fisherman' (Apte. Lexicon) Such people are referred to in Rgveda by Viśvāmitra as 'Bhāratam janam.' Mahābhārata alludes to 'thousands of mlecchas', a numerical superiority equaled by their valour and courage in battle which enhances the invincibility of Pandava (MBh. 7.69.30; 95.36).

Excerpt from Encyclopaedia Iranica article on cognate *dahyu* country (often with reference to the people inhabiting it): DAHYU (OIr. *dahyu-*), attested in Avestan *daẋiiu-, daŋhu-* "country" (often with reference to the people inhabiting it; cf. *AirWb.*, cot. 706; Hoffmann, pp. 599-600 n. 14; idem and Narten, pp. 54-55) and in Old Persian *dahyu-* "country, province" (pl. "nations"; Gershevitch, p. 160). The term is likely to be connected with Old Indian *dásyu* "enemy" (of the Aryans), which acquired the meaning of "demon, enemy of the gods" (Mayrhofer, *Dictionary* II, pp. 28-29). Because of the Indo-Iranian parallel, the word may be traced back to the root *das-*, from which a term denoting a large collectivity of men and women could have been derived. Such traces can be found in Iranian languages: for instance, in the ethnonym Dahae (q.v., i) "men" (cf. Av. ethnic name [fem. adj.] *dāhī*, from *dåŋha-*; *AirWb.*, col. 744; Gk. Dáai, etc.), in Old Persian *dahā* "the Daha people" (Brandenstein and Mayrhofer, pp. 113-14), and in Khotanese *daha* "man, male" (Bailey, *Dictionary*, p. 155).

In Avestan the term did not have the same technical meaning as in Old Persian. Avestan *daẋiiu-, daŋhu-* refers to the largest unit in the vertical social organization. See, for example, Avestan *xᵛaētu-* (in the Gathas) "next of kin group" and *nmāna-* "house," corresponding to Old Persian *taumā-* "family"; Avestan *vīs-* "village," corresponding to Avestan *vərəzəna-* "clan"; Avestan *zantu-* "district"; and Avestan *daẋiiu-, daŋhu-* (Benveniste, 1932; idem, 1938, pp. 6, 13; Thieme, pp. 79ff.; Frye, p. 52; Boyce, *Zoroastrianism* I, p. 13; Schwartz, p. 649; Gnoli, pp. 15ff.). The connection *daẋiiu, daŋhu-* and *arya-* "Aryans" is very common to indicate the Aryan lands and peoples, in some instances in the plural: *airiiå daŋhāuuō, airiianąm daẋiiunąm, airiiābiiō daŋhubiiō*. In *Yašt* 13.125 and 13.127 five countries (*daẋiiu-*) are mentioned, though their identification is unknown or uncertain; in the same *Yašt* (13.143-44) the countries of other peoples are added to those of the Aryans: *tūiriia, sairima, sāinu, dāha*.

In Achaemenid inscriptions Old Persian *dahyu-* means "satrapy" (on the problems relative to the different lists of *dahyāva* [pl.], cf. Leuze; Junge; Walser, pp. 27ff.; Herzfeld, pp. 228-29; Herrenschmidt, pp. 53ff.; Calmeyer, 1982, pp. 105ff.; idem, 1983, pp. 141ff.) and "district" (e.g., Nisāya in Media; DB 1.58; Kent, *Old Persian*, p. 118). The technical connotation of Old Persian *dahyu* is certain and is confirmed—despite some doubts expressed by George Cameron but refuted by Ilya Gershevitch—by the loanword *da-a-yau-iš* in Elamite. On the basis of the hypothetical reconstruction of twelve "districts" and twenty-nine "satrapies," it has been suggested that the formal identification of the Old Persian numeral 41 with the ideogram *DH*, sometimes used for *dahyu* (Kent, *Old Persian*, pp. 18-19), can be explained by the fact that there were exactly forty-one *dahyāva* when the sign *DH* was created (Mancini).

From the meaning of Old Persian *dahyu* as "limited territory" come Middle Persian and Pahlavi *deh* "country, land, village," written with the ideogram *MTA* (*Frahang ī Pahlawīg* 2.3, p. 117; cf.

Syr. *mātā*), and Manichean Middle Persian *dyh*(MacKenzie, p. 26). At times the Avestan use is reflected in Pahlavi *deh*, but already in Middle Persian the meaning "village" is well documented; it appears again in Persian *deh*.

That Pali uses the term 'milakkhu' is significant (cf. *Uttarādhyayana Sūtra* 10.16) and reinforces the concordance between 'mleccha' and 'milakkhu' (a pronunciation variant) and links the language with 'meluhha' as a reference to a language in Mesopotamian texts and in the cylinder seal of Shu-ilishu. [Possehl, Gregory, 2006, Shu-ilishu's cylinder seal, Expedition, Vol. 48, No. 1 http://www.penn.museum/documents/publications/expedition/PDFs/48-1/What%20in%20the%20World.pdf] This seal shows a sea-faring Meluhha merchant who needed a translator to translate meluhha speech into Akkadian. The translator's name was Shu-ilishu as recorded in cuneiform script on the seal. This evidence rules out Akkadian as the Indus or Meluhha language and justifies the search for the proto-Indian speech from the region of the Sarasvati river basin which accounts for 80% (about 2000) archaeological sites of the civilization, including sites which have yielded inscribed objects such as Lothal, Dwaraka, Kanmer, Dholavira, Surkotada, Kalibangan, Farmana, Bhirrana, Kunal, Banawali, Chandigarh, Rupar, Rakhigarhi. The language-speakers in this basin are likely to have retained cultural memories of Indus language which can be gleaned from the semantic clusters of glosses of the ancient versions of their current *lingua franca* available in comparative lexicons and nighaṇṭu-s.

**Evidence from Valmiki Rāmāyaṇa**

Slokas 5.30.16 to 21 in the 29[th] sarga of Sundara Kandam, provide an episode of Hanuman introspecting on the language in which he should speak to Sita. This evidence refers to two dialects:
Sanskrit and mānuṣam vākyam (lit. jāti bhāṣā). In this narrative mānuṣam vākyam (spoken dialect) is distinguished from Sanskrit of a Brahmin (or, grammatically correct and well-prouncedd Sanskrit used in yajña-s).

*1. "antaramtvaha māsādya rākṣasīnam iha sthitah"*

*2. "śanairāśvāsaiṣyāmi santāpa bahulām imām"*

(Staying here itself and getting hold of an opportunity even in the midst of the female-demons (when they are in attentive), I shall slowly console Sita who is very much in distress. )

*3. "aham hi atitanuścaiva vānara śca viśeṣata"*

*4. "vācam ca udāhariṣyāmi mānuṣīm iha samskṛtām"*
(However, I am very small in stature, particularly as a monkey and can speak now Sanskrit, the human language too.)

*5. "yadi vācam pradāsyami dwijātiriva samskṛtām"*

*6. "rāvaṇam manyamānā mām sītā bhītā bhavi ṣyati"*

*7. vānarasya viśeṣena kathamsyādabibhāṣaṇam*
(If I use Sanskrit language like a llsde, Sita will get frightened, thinking that Rāva ṇ a has come disguised as a monkey. Especially, how can a monkey speak it?)

*8. "avaśyameva vaktavyam mānuṣam vākyam arthavat"*

*9. "mayā śāntvayitum śakyā"*

*10. "nānyathā iyam aninditā"*
(Certainly, meaningful words of a human being are to be spoken by me. Otherwise, the virtuous Sita cannot be consoled.)

*11. "sā iyam ālokya me rūpam jānakī bhāṣitam tathā ||*

*rakṣobhih trāsitaa pūrvam bhuūah trūsam gamiṣyati |"*

(Looking at my figure and the language, Seetha who was already frightened previously by the demons, will get frightened again.) [Translation based on http://www.valmikiramayan.net/sundara/sarga30/sundara_30_frame.htm See: Narayana Iyengar, 1938, Vanmeegarum Thamizhum; http://tashindu.blogspot.com/2006_12_01_archive.html In this work, Narayana Iyengar cites that the commentator interpret mānuṣam vākyam as the language spoken in Kosala.]

Evidence from Śatapatha Brāhmaṇa for *mleccha vācas*

An extraordinary narrative account from Śatapatha Brāhmaṇa is cited in full to provide the context of the yagna in which vaak (speech personified as woman) is referred to the importance of grammatical speech in yagna performance and this grammatical, intelligible speech is distinguished from mlecccha, unintelligible speech. The example of the usage of phrase 'he 'lavo is explained by Sayana as a pronunciation variant of: 'he 'rayo. i.e. 'ho, the spiteful (enemies)!' This grammatically correct phrase, the Asuras were unable to pronounce correctly, notes Sayana. The ŚB text and translation are cited in full because of the early evidence provided of the mleccha speech (exemplifying what is referred to Indian language studies as 'ralayo rabhedhah'; the transformed use of 'la' where the syllable 'ra' was intended. This is the clearest evidence of a proto-Indian language which had dialectical variants in the usage by asuras and devas (i.e. those who do not perform yagna and those who perform yagna using vaak, speech.) This is comparable to mleccha vācas and ārya vācas differentiation by Manu. The text of ŚB 3.2.1.22-28 and translation are as follows:

*yoṣā vā iyaṃ vāgyadenaṃ na yuvitehaiva mā tiṣṭhantamabhyehīti brūhi tāṃ tu na āgatāṃ pratiprabrūtāditi sā hainaṃ tadeva tiṣṭhantamabhyeyāya tasmādu strī pumāṃsaṃ saṃskṛte tiṣṭhantamabhyaiti tāṃ haibhya āgatāṃ pratiprovāceyaṃ vā āgāditi tāṃ devāḥ |*

*asurebhyo 'ntarāyaṃstāṃ svīkṛtyāgnāveva parigṛhya sarvahutamajuhavurāhutirhi devānāṃ sa yāmevāmūmanuṣṭubhājuhavustadevainām taddevāḥ svyakurvata te 'surā āttavacaso he 'lavo he 'lava iti vadantaḥ parābabhūvuḥ atraitāmapi vācamūduḥ |*

*upajijñāsyāṃ sa mlecastasmānna brāhmaṇo mlecedasuryā haiṣā vā natevaiṣa dviṣatāṃ sapatnānāmādatte vācaṃ te 'syāttavacasaḥ parābhavanti ya evametadveda o 'yaṃ yajño vācamabhidadhyau |*

*mithunyenayā syāmiti tāṃ sambabhūva indro ha vā īkṣāṃ cakre |*

*mahadvā ito 'bhvaṃ janiṣyate yajñasya ca mithunādvācaśca yanmā tannābhibhavediti sa indra eva garbho bhūtvaitanmithunam praviveśa sa ha saṃvatsare jāyamāna īkṣāṃ cakre |*

*mahāvīryā vā iyaṃ yoniryā māmadīdharata yadvai meto mahadevābhvaṃ nānuprajāyeta yanmā tannābhibhavediti tāṃ pratiparāmṛśyaveṣṭyācinat |*

*tāṃ yajñasya śīrṣanpratyadadhādyajño hi kṛṣṇaḥ sa yaḥ sa yajñastatkṛṣṇājinaṃ yo sā yoniḥ sā kṛṣṇaviṣāṇātha yadenāmindra āveṣṭyācinattasmādāveṣṭiteva sa yathaivāta indro 'jāyata garbho bhūtvaitasmānmithunādevamevaiṣo 'to jāyate garbho bhūtvaitasmānmithunāt tāṃ vā uttānāmiva badhnāti |*

Translation: 22.The gods reflected, 'That Vaak being a woman, we must take care lest she should allure him. – Say to her, "Come hither to make me where I stand!" and report to us her having come.' She then went up to where he was standing. Hence a woman goes to a man who stays in a well-trimmed (house). He reported to them her having come, saying, 'She has indeed come.' 23. The gods then cut her off from the Asuras; and having gained possession of her and enveloped her completely in fire, they offered her up as a holocaust, it being an offering of the gods. (78) And in that they offered her with an anushtubh verse, thereby they made her their own; and the Asuras being deprived of speech, were undone, crying, 'He 'lavah! He 'lavah!' (79) 24. Such was the unintelligible speech which they then uttered, -- and he (who speaks thus) is a Mlekkha (barbarian). Hence let no Brahman speak barbarous language, since such is the speech of the Asuras. Thus alone he deprives his spiteful enemies of speech; and whosoever knows this, his enemies, being deprived of speech, are undone. 25. That Yajna (sacrifice) lusted after Vaak (speech [80]), thinking, 'May I pair with her!' He united with her. 26. Indra then thought within himself, 'Surely a great monster will spring from this union of Yagna and Vaak: [I must take care] lest it should get the better of me.' Indra himself then became an embryo and entered into that union. 27. Now when he was born after a year's time, he thought within himself, 'Verily of great vigour is this womb which has contained me: [I must take care] that no great monster shall be born from it after me, lest it should get the better of me!' 28. Having seized and pressed it tightly, he tore it off and put it on the head of Yagna (sacrifice [81]); for the black (antelope) is the sacrifice: the black deer skin is the same as that sacrifice, and the black deer's horn is the same as that womb.  And because it was by pressing it tightly together that Indra tore out (the womb), therefore it (the horn) is bound tightly (to the end of the garment); and as Indra, having become an embryo, sprang from that union, so is he (the sacrifice), after becoming an embryo, born from that union (of the skin and the horn). (ŚB 3.2.1.23-25). (fn 78) According to Sayana, 'he 'lavo' stands for 'he 'rayo' (i.e. ho, the spiteful (enemies)!' which the Asuras were unable to pronounce correctly. The Kaanva text, however, reads te hātavāko 'su  hailo haila ity etām ha vācam vadantah parābabhūvuh (? i.e. he p. 32 ilaa, 'ho, speech'.) A third version of this passage seems to be referred to in the Mahā  bhāṣya (Kielh.), p.2. (p.38). (fn 79) Compare the corresponding legend about Yagna and Dakṣiṇā  (priests' fee), (Taitt. S. VI.1.3.6. (p.38) (fn 79) 'Yagnasya sīrṣan'; one would expect 'kṛṣṇa(sāra)sya sīrṣan.' The Taitt.S. reads 'tām mṛgeṣu ny adadhāt.' (p.38) (fn81) In the Kanva text 'atah (therewith)' refers to the head of the sacrifice, -- sa yak khirasta upasprisaty ato vā enām etad agre pravisan pravisaty ato vā agre gāyamāno gāyate tasmāk khirasta upasprisati. (p.39)(cf. śatapatha Brāhmaṇa vol. 2 of 5, tr. By Julius Eggeling, 1885, in SBE Part 12; fn 78-81).

Mesopotamian texts refer to a language called meluhha (which required an Akkadian translator); this meluhha is cognate with mleccha. Seafaring meluhhan merchants used the script in trade transactions; artisans created metal artifacts, lapidary artificats of terracotta, ivory for trade. Glosses of the proto-Indic or Indus language are used to read rebus the Indus script inscriptions. The glyphs of the script include both pictorial motifs and signs and both categories of glyphs are read rebus. As a first step in delineating the Indus language, an Indian lexicon provides a resource, compiled semantically cluster over 1240 groups of glosses from ancient Indian languages as a proto-Indic substrate dictionary.
See http://www.scribd.com/doc/2232617/lexicon linked at http://sites.google.com/site/kalyan97/indus-writing

"The word *meluh.h.a* is of special interest. It occurs as a verb in a different form (mlecha-) in Vedic only in ŚB 3.2.1, an eastern text of N. Bihar where it indicates 'to speak in barbarian fashion'. But it has a form closer to Meluh.h.a in Middle Indian (MIA): Pali, the church language of S. Buddhism which originated as a western N. Indian dialect (roughly, between Mathura, Gujarat and the Vindhya) has milakkha, milakkhu. Other forms, closer to ŚB mleccha are found in MIA *mliccha > Sindhi milis, Panjabi milech, malech, Kashmiri bri.c.hun 'weep, lament' (< *mrech-, with the common r/l interchange of IA), W. Pahari mel+c.h 'dirty'. It seems that, just as in other cases mentioned above, the original local form *m(e)luh. (i.e. m(e)lukh in IA pronunciation, cf. E. Iranian bAxdhI 'Bactria' > AV *bahli-ka, balhi-ka) was preserved only in the South (Gujarat? >Pali), while the North (Panjab, Kashmir, even ŚB and Bengal) has *mlecch. The sound shift from -h.h.-/-kh- > -cch- is unexplained; it may have been modeled on similar correspondences in MIA (Skt. Akṣi 'eye' _ MIA akkhi, acchi; ks.Etra '_eld' _ MIA khetta, chetta, etc.) The meaning of Mleccha must have evolved from 'self-designation' > 'name of foreigners', cf. those of the Franks > Arab farinjI 'foreigner.' Its introduction into Vedic must have begun in Meluh.h.a, in Baluchistan-Sindh, and have been transmitted for a long time in a non-literary level of IA as a nickname, before surfacing in E. North India in Middle/Late Vedic as Mleccha. (Pali milāca is influenced by a `tribal' name, Piśā ca, as is Sindhi milindu, milidu by Pulinda; the word has been further `abbreviated' by avoiding the difficult cluster ml- : Prākṛt mecha, miccha, Kashmiri m ĭ c(h), Bengali mech (a Tib.-Burm tribe) and perhaps Pashai mece if not < *mēcca `defective' (Turner, CDIAL 10389. | Parpola 1994: 174 has attempted a Dravidian explanation. He understands Meluh.h. a (var. Melah.h.a) as Drav. *Mēlakam [mēlaxam] `high country' (= Baluchistan) (=Ta-milakam) and points to Neo-Assyr. Baluh.h.u `galbanum', sinda `wood from Sindh'. He traces mlech, milakkha back to *mleks. , which is seen as agreeing, with central Drav. Metathesis with *mlēxa = mēlaxa-m. Kuiper 1991:24 indicates not infrequent elision of (Dravid.) —a- when taken over into Skt. | Shafer 1954 has a Tib-Burm. Etymology *mltse; Southworth 1990: 223 reconstructs Pdrav. 2 *muzi/mizi `say, speak, utter', DEDR 4989, tamil `Tamil' < `own speech'.)" [Witzel, Michael, 1999, Substrate Languages in Old Indo-Aryan (Rgvedic, Middle and Late Vedic, *Electronic Journal of Vedic Studies* (EJVS) 5-1 (1999) pp.1-67. http://www.ejvs.laurasianacademy.com/ejvs0501/ejvs0501article.pdf]

Note: Coining a term, "Para-Munda", denoting a hypothetical language related but not ancestral to modern Munda languages, the author goes on to identify it as "Harappan", the language of the Harappan civilization. The author later recounts this and posits that Harappan were illiterate and takes the glyphs of the script to be symbols without any basis in any underlying language.[cf. Steve Farmer, Richard Sproat, and Michael Witzel, 2005, The Collapse of the Indus-Script Thesis: The Myth of a Literate Harappan Civilization, EJVS 11-2 Dec. 13, 2005.]

ṛgveda (ṛca 3.53.12) uses the term, *'bhāratam janam'*, which can be interpreted as 'bhārata folk'. The ṛṣi of the sūkta is viśvāmitra gāthina. India was called Bhāratavarṣa after the king Bhārata. (Vāyu 33, 51-2; Bd. 2,14,60-2; lin:ga 1,47,20,24; Viṣṇu 2,1,28,32).

*Ya ime rodasī ubhe aham indram atuṣṭavam*

*viśvāmitrasya rakṣati brahmedam bhāratam janam*

3.053.12 I have made Indra glorified by these two, heaven and earth, and this prayer of viśvāmitra protects the people of Bhārata. [Made Indra glorified: indram atuṣṭavam — the verb is the third preterite of the casual, I have caused to be praised; it may mean: I praise Indra, abiding between heaven and earth, i.e. in the firmament].

The evidence is remarkable that almost every single glyph or glyptic element of the Indus script can be read rebus using the repertoire of artisans (lapidaries working with precious shell, ivory, stones and terracotta, mine-workers, metal-smiths working with a variety of minerals, furnaces and other tools) who created the inscribed objects and used many of them to authenticate their trade transactions. Many of the inscribed objects are seen to be calling cards of the professional artisans, listing their professional skills and repertoire.

The identification of glosses from the present-day languages of India on Sarasvati river basin is justified by the continuation of culture evidenced by many artifacts evidencing civilization continuum from the Vedic Sarasvati River basin, since language and culture are intertwined, continuing legacies:

Huntington notes [http://huntingtonarchive.osu.edu/Makara%20Site/makara]: "There is a continuity of composite creatures demonstrable in Indic culture since Kot Diji ca. 4000 BCE."

Mriga (pair of deer or antelope) in Buddha sculptures compare with Harappan period prototype of a pair of ibexes on the platform below a seated yogin. http://tinyurl.com/gonsh

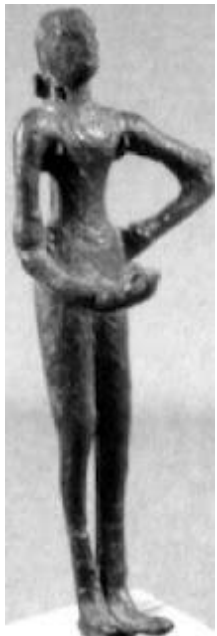

Continued use of śankha (turbinella pyrum) bangles which tradition began 6500 BCE at Nausharo;

Continued wearing of sindhur at the parting of the hair by married ladies as evidenced by two terracotta toys painted black on the hair, painted golden on the jewelry and painted red to show sindhur at the parting of the hair;

Finds of shivalinga in situ in a worshipful state in Harappa (a metaphor of Mt. Kailas summit where Maheśvara is in tapas, according to Hindu tradition);

Terracotta toys of Harappa and Mohenjo-daro showing Namaste postures and yogasana postures;

Three-ring ear-cleaning device

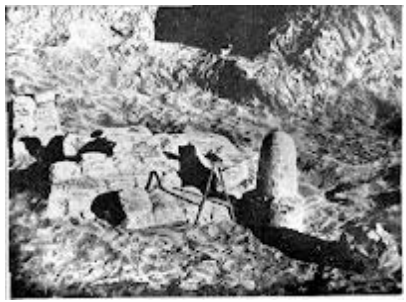

Legacy of architectural forms

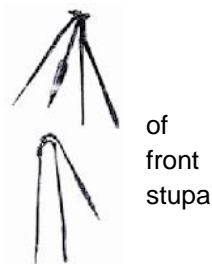

Legacy of puṣkariṇi in front mandirams; as in of Mohenjo-daro

of front stupa

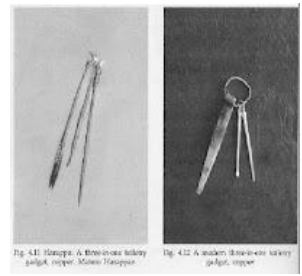

Legacy of metallurgy and the writing system on punch-marked coins

Legacy of continued use of cire perdue technique for making utsava bera (bronze murti)

Legacy: Engraved celt tool of Sembiyan-kandiyur with Sarasvati hieroglyphs: calling-card of an artisan

Legacy of acharya wearing uttariyam (shawl) leaving right-shoulder bare

Form of addressing a person respectfully as: arya, ayya (Ravana is also referred to as arya in the Great Epic Rāmāyaṇa)

Plate X [c] Lingam in situ in Trench Ai (MS Vats, 1940, Excavations at Harappa, Vol. II, Calcutta) Lingam, grey sandstone in situ, Harappa, Trench Ai, Mound F, Pl. X (c) (After Vats). "In an earthenware jar, No. 12414, recovered from Mound F, Trench IV, Square I… in this jar, six lingams were found along with some tiny pieces of shell, a unicorn seal, an oblong grey sandstone block with polished surface, five stone pestles, a stone palette, and a block of chalcedony…" (Vats, MS, 1940, Excavations at Harappa, Delhi, p. 370).

Continued use of cire perdue technique of bronze-casting. Bronze murti: cire perdue technique used today in Swamimalai to make bronze utsavabera (idols carried in procession). Eraka Subrahmanya is the presiding divinity in Swamimalai. Eraka! Copper.Devices on punch-marked coins comparable to Sarasvati hieroglyphs.

Toilet gadgets: Ur and Harappa After Woolley 1934, Vats 1941

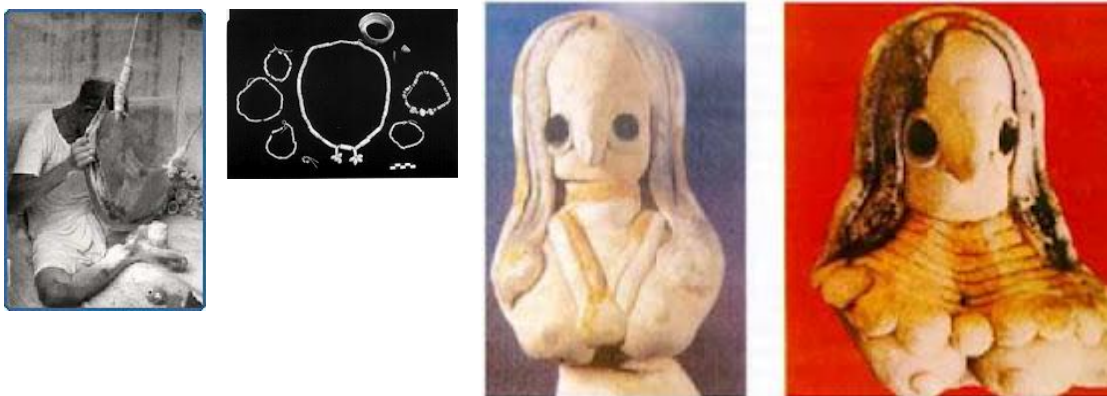

Nausharo: female figurines. Wearing sindhur at the parting of the hair. Hair painted black, ornaments golden and sindhur red. Period 1B, 2800 – 2600 BCE. 11.6 x 30.9 cm.[After Fig. 2.19, Kenoyer, 1998].

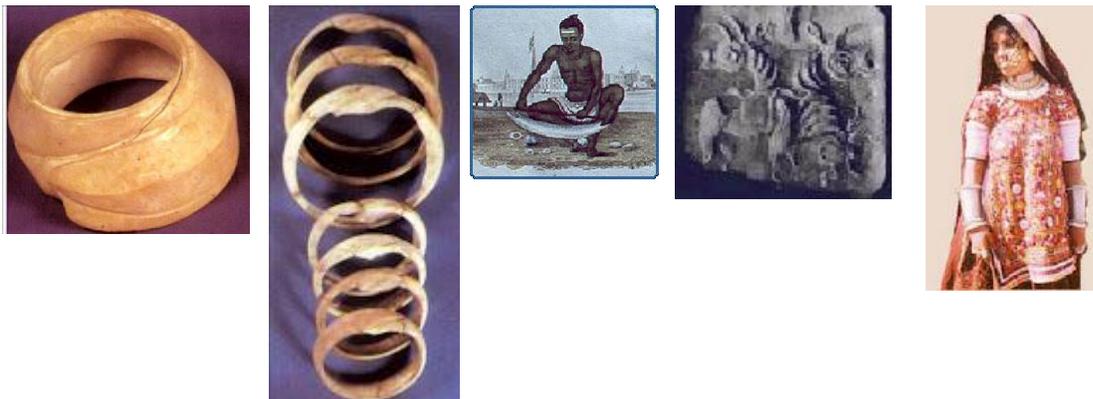

Śankha artifacts: Wide bangle made from a single conch shell and carved with a chevron motif, Harappa; marine shell, Turbinella pyrum (After Fig. 7.44, Kenoyer, 1998) National Museum, Karachi. 54.3554. HM 13828. Seal, Bet Dwaraka 20 x 18 mm of conch shell. Seven shell bangles from burial of an elderly woman, Harappa; worn on the left arm; three on the upper arm and four on the forearm; 6.3 X 5.7 cm to 8x9 cm marine shell, Turbinella pyrum (After Fig. 7.43, Kenoyer, 1998) Harappa museum. H87-635 to 637; 676 to 679. Modern lady from Kutch, wearing shell-bangles.

6500 BCE. Date of the woman's burial with ornaments including a wide bangle of shankha. Mehergarh. Burial ornaments made of shell and stone disc beads, and turbinella pyrum (sacred conch, śankha) bangle, Tomb MR3T.21, Mehrgarh, Period 1A, ca. 6500 BCE. The nearest source for this shell is Makran coast near Karachi, 500 km. South. [After Fig. 2.10 in Kenoyer, 1998]. Śankha wide bangle and other ornaments, c. 6500 BCE (burial of a woman at Nausharo). Glyph: 'shell-cutter's saw'

Some miniature tablets with Indus inscriptions are shaped like a shell-cutter's saw shown in the photograph of a bangle-maker from Bengal, cutting *turbinella pyrum*. Shapes of some text glyphs also resemble the shell-cutter's saw:

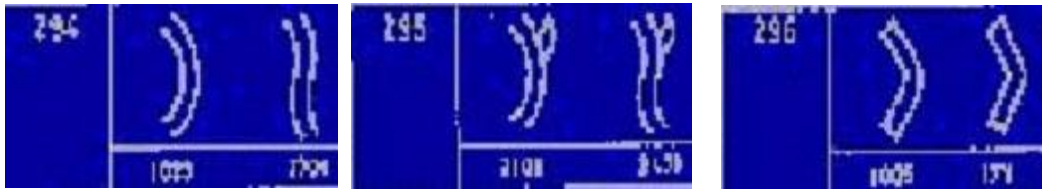

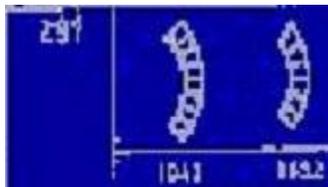

V294 V295 V296 V297

It is unlikely that Akkadian was a possible underlying language because a cuneiform cylinder seal with an Akkadian inscription, showing a seafaring Meluhhan merchant (carrying an antelope) required an interpreter, Shu-ilishu, confirming that the Meluhhan's language was not Akkadian. There is substantial agreement among scholars pointing to the Indian civilization area as a linguistic area.

I suggest that Meluhha mentioned in Mesopotamian texts of 3rd-2nd millennium BCE is a language of this linguistic area. That meluhha and mleccha are cognate and that mleccha is attested as a mleccha vācas (mleccha speech) distinguished from arya vācas (arya speech) indicates that the linguistic area had a colloquial, ungrammatical mleccha speech – *lingua franca* and a grammatically correct arya speech – literary language. The substrate glosses of the Indian lexicon are thus reasonably assumed to be the glosses of mleccha vācas, the speech of the artisans who produced the artifacts and the inscribed objects with the writing system. This assumption is further reinforced by the fact that about 80% of archaeological sites of the civilization are found on the banks of Vedic River Sarasvati leading some scholars to rename the Indus Valley civilization as Sarasvati-Sindhu civilization.

In this context, the following monumental work by Sylvan Levi, Jules Bloch and Jean Przyluski published in the 1920's continues to be relevant, even today, despite some advances in studies related to formation of Indian languages and the archaeological perspectives of and evidences from the civilization.

Przyluski notes the principal forms of the words signifying 'man' and 'woman' in the Munda languages:

Man: hor, hōrol, harr, hŏr, haṛa, hoṛ, koro

Woman: kūṛĩ, ērā, koṛi, kol

Comparing 'son' and 'daughter' in Santali:

Son = kora hapan; daughter = kuri hapan

"…a root kur, kor is differentiated in the Munda languages for signifying: man, woman, girl and boy. That in some cases this root has taken a relatively abstract sense is proved by Santali koḍa, koṛa, which signify 'one' as in the expression 'koḍa ke koḍa' 'each single one'. Thus one can easily understand that the same root has served the purpose of designating the individual not as an indivisible unity but as a numerical whole…Thus we can explain the analogy between the root kur, kor 'man' the number 20 in Munda kūṛī kūṛī , koḍī and the number 10 in Austro-Asiatic family ko, se-kūr, skall, gal." (ibid., pp. 28-30).

Homonym: कोल [ kōla ] *n* An income, or goods and chattels, or produce of fields &c. seized and sequestered (in payment of a debt). *V* धरून ठेव, सोड. 2 *f* The hole dug at the game of विटीदांडू, at marbles &c. कोलणें [ kōlaṇēṃ ] *v c* To strike the विटी in the hole कोली with the bat or दांडू. (In the game of विटीदांडू) 2 To cast off from one's self upon another (a work). Ex. पैका मागावयास लागलों म्हणजे बाप लेंकावर कोल-तो लेंक बापावर कोलतो. 3 To cast aside, reject, disallow, flout, scout. कोलून मारणें To kick up the heels of; to trip up: also to turn over (from one side to the other). किरकोळी [ kirakōḷī ] *f* (किरकोळ) A heap of miscellaneous articles.

An old Munda word, kol means 'man'. S. K. Chatterjee called the Munda family of languages as Kol, as the word, according to him, is (in the Sanskrit-Prākṛt form Kolia) an early Aryan modification of an old Munda word meaning 'man'. [Chatterjee, SK, The study of kol, *Calcutta Review*, 1923, p. 455.] Przyluski accepts this explanation. [Przyluski, Non-aryan loans in Indo-Aryan, in: Bagchi, PC, *Pre-aryan and pre-dravidian,* pp.28-29 http://www.scribd.com/doc/33670494/prearyanandpredr035083mbp]

Note: This area can be called speakers of 'mleccha, meluhha' or mleccha vācas according to Manusmṛti (lingua franca of the artisans). Manusmṛti distinguishes two spoken language-groups: mleccha vācas and arya vaacas (that is, spoken dialect distinguished from grammatically correct glosses).

"A *Sprachbund*…in German, plural "Sprachbünde" IPA, from the German word for "language union", also known as a linguistic area, convergence area, or diffusion area, is a group of languages that have become similar in some way because of geographical proximity and language contact. They may be genetically unrelated, or only distantly related. Where genetic affiliations are unclear, the *sprachbund* characteristics might give a false appearance of relatedness…In a classic 1956 paper titled "India as a Linguistic Area", Murray Emeneau [Emeneau, Murray. 1956. India as a Lingusitic Area. "Langauge" 32: 3-16. http://en.academic.ru/dic.nsf/enwiki/113093] laid the groundwork for the general acceptance of the concept of a *Sprachbund*. In the paper, Emeneau observed that the subcontinent's Dravidian and Indo-Aryan languages shared a number of features that were not inherited from a common source, but were areal features, the result of diffusion during sustained contact." Common features of a group of languages in a *Sprachbund* are called 'areal features'. In linguistics, an areal feature is any typological feature shared by languages within the same geographical area. An example refers to retroflex consonants in the Burushaski {Berger, H. Die Burushaski-Sprache von Hunza und Nagar. Vols. I-III. Wiesbaden: Harrassowitz 1988 ] [Tikkanen (2005)]}, Nuristani [G.Morgenstierne, Irano-Dardica. Wiesbaden 1973], Dravidian, Munda and Indo-Aryan language families of the Indian subcontinent. *The Munda Languages*. Edited by Gregory D. S. Anderson. London and New York: Routledge (Routledge Language Family Series), 2008.

Notes on Indian linguistic area: pre-aryan,pre-Munda and pre-dravidian in India

It will be a hasty claim to make that Old Tamil or Proto-Munda or Santali or Prakṛt or Pali or any other specific language of the Indian linguistic area, by itself (to the exclusion of other languages in contact), explains the language of the Indus civilization. In this context, the work by Sylvan Levi, Jules Bloch and Jean Przyluski published in the 1920's (cited elsewhere) continues to be relevant, even today, despite some advances in studies related to formation of Indian languages and the archaeological perspectives of and evidences from the civilization.

Some glyphs of the script are yet to be decoded. Tentative readings of such glyphs yet to be validated by the cipher code key of Indus script are detailed (including decipherment of inscriptions from scores of small sites) at http://sites.google.com/site/kalyan97/induswriting If the glyphs are unambiguously identified and read in archaeological context and the context of other glyphs of the inscription itself, it will be possible to decipher them. For this purpose, some graphemes (which have homonyms and can be read rebus) are provided from the Indian Lexicon of the Indian linguistic area.

Graphemes:

kola 'tiger' (Telugu); 'jackal' (Konkani); kul id. (Santali)

kol 'the name of a bird, the Indian cuckoo' (Santali)

kolo 'a large jungle climber, dioscorea doemonum (Santali)

kulai 'a hare' (Santali)

Grapheme: Ta. kōl stick, staff, branch, arrow. Ma. kōl staff, rod, stick, arrow. Ko. kl stick, story of funeral car. To. kwṣ stick. Ka. kōl, kōlu stick, staff, arrow. Koḍ. Klï stick. Tu. kōlǔ, kōlustick, staff. Te. kōla id., arrow; long, oblong; kōlana elongatedness, longation; kōlani elongated. Kol. (SR.) kolā, (Kin.) kōla stick. Nk. (Ch.) kōl pestle. Pa. kōl shaft of arrow.Go. (A.) kōla id.; kōlā (Tr.) a thin twig or stick, esp. for kindling a fire, (W. Ph.) stick, rod, a blade of grass, straw; (G. Mu. Ma. Ko.) kōla handle of plough, sickle, knife, etc. ( Voc.988); (ASu.) kōlā stick, arrow, slate-pencil; (LuS.) kola the handle of an implement. Konḍa kōl big wooden pestle. Pe. kōlpestle. Manḍ. kūl id. Kui kōḍu (pl. kōṭka) id. Kuwi (F.)kōlū (pl. kōlka), (S. Su.) kōlu (pl. kōlka) id. Cf. 2240 Ta.kōlam (Tu. Te. Go.). / Cf. OMar. (Master) kōla stick. (DEDR 2237). कोलदंडा or कोलदांडा [ kōladaṇḍā or kōladāṇḍā ] m A stick or bar fastened to the neck of a surly dog. (Marathi)

kola [ kōla ] f. The bandicoot rat, mus malibaricos (Rajasthani)

Skanda Purana refers to kol as a mleccha community. (Hindu *śabdasagara*).

kolhe, 'the koles, are an aboriginal tribe of iron smelters speaking a language akin to that of Santals' (Santali) kōla m. name of a degraded tribe Hariv. Pk. Kōla — m.; B. kol name of a Muṇḍā tribe (CDIAL 3532). A Bengali lexeme confirms this: কোল[1] [ kōla[1] ] an aboriginal tribe of India; a member of this tribe. (Bengali) That in an early form of Indian linguistic area, kol means 'man' gets substantiated by a Nahali and Assamese glosses: kola 'woman'. See also: Wpah. Khaś.kuṛi, cur. kuḷī, cam. kŏḷā ' boy ', Sant. Muṇḍari koṛa ' boy ', kuṛi ' girl ', Ho koa, kui, Kūrkū kōn, kōnjē). Prob. separate from RV. kr̥tā -- ' girl ' H. W. Bailey TPS 1955, 65; K. kūrü f. ' young girl ', kash. kōṛī, ram. kuṛhī; L. kuṛā m. ' bridegroom ', kuṛī f. ' girl, virgin, bride ', awāṇ. kuṛī f. ' woman '; P. kuṛī f. ' girl, daughter ', (CDIAL 3295). कारकोळी or ळ्या [ kārakōḷī or ḷyā ] a Relating to the country कार- कोळ--a tribe of Bráhmans (Marathi).

**Mleccha and Bharatiya languages**

Mleccha was substratum language of bharatiyo (casters of metal) many of whom lived in dvīpa (land between two rivers –Sindhu and Sarasvati -- or islands on Gulf of Kutch, Gulf of Khambat, Makran coast and along the Persian Gulf region of Meluhha).

Mleccha were bharatiya (Indians) of Indian linguistic area

According to Matsya Purāṇa (10.7), King Veṇa was the ancestor of the mleccha; according to Mahābhārata (MB. 12.59, 101-3), King Veṇa was a progenitor of the Niṣāda dwelling in the Vindhya mountains. Nirukta 3.8 includes Niṣāda among the five peoples mentioned in the r̥gveda 10.53.4, citing Aupamanyava; the five peoples are: brāhmaṇa, kṣatriya, vaiśya, śūdra and Niṣāda. Niṣāda gotra is mentioned in the gaṇapāṭha of Pāṇini (Aṣṭādhyāyī 4.1.100). Niṣāda were mleccha. It should be noted that Pāṇini associated yavana with the Kāmboja (Pāṇini, *Gaṇapāṭha*, 178 on 2.1.72).

Mullaippāṭṭu (59-66) (composed by kāvirippūmpāṭṭinattuppon vāṇigaṇār mahanāṛṇ.appūḍanār) are part of Pattuppāṭṭu, ten Tamil verses of Sangam literature; these refer to a chief of Tamil warriors whose battle-field tent was built by Yavana and guarded by mleccha who spoke only through gestures. (JV Chelliah, 1946, *Pattuppāṭṭu; ten Tamil idylls, translated into English verse*, South India Saiva Siddhanta Works Publishing Society, p. 91).

Mahābhārata notes that the Pāṇḍava army was protected by mleccha, among other people (Kāmboja, śaka, Khasa, Salwa, Matsya, Kuru, Mleccha, Pulinda, Draviḍa, Andhra and Kāñci) (MBh. V.158.20). Sūta laments the misfortune of the Kaurava-s: 'When the Nārāyaṇa-s have been killed, as also the Gopāla-s, those troops that were invincible in battle, and many thousands of mleccha-s, what can it be but Destiny?' (MBh. IX.2.36: *Nārāyaṇā hatāyatra Gopālā yuddhadurmahāḥ mlecchāśca bahusāhasrāh kim anyad bhāgadheyatah*?)

**Nahali, Meluhhan, Language 'X'**

On the banks of River Narmada are found speakers of Nahali, the so-called language isolate with words from Indo-Aryan, Dravidian and Munda – which together constitute the indic language substratum of a linguistic area, ca. 3300 BCE on the banks of Rivers Sarasvati and Sindhu – a region referred to as Meluhha in Mesopotamian cuneiform records; hence the language of the inscribed objects can rightly be called Meluhhan or Mleccha, a language which Vidura and Yudhiṣṭhira knew (as stated in the Great Epic, Mahābhārata).Elsewhere in the Great Epic we read how Sahadeva, the youngest of the Pāṇḍava brothers, continued his march of conquest till he reached several islands in the sea (no doubt with the help of ships) and subjugated the Mleccha inhabitants thereof. Brahmāṇḍa 2.74.11, Brahma 13.152, Harivaṁśa 1841, Matsya 48.9, Vāyu 99.11, cf. also Viṣṇu 4.17.5, Bhāgavata 9.23.15, see Kirfel 1927: 522: *pracetasah putraśatam rājanah sarva eva te // mleccharāṣṭrādhipāh sarve udīcīm diśam āśritāh* which means, of course, not that these '100' kings conquered the 'northern countries' way beyond the Hindukuṣ or Himalayas, but that all these 100 kings, sons of pracetās (a descendant of a 'druhyu'), kings of mleccha kingdoms, are 'adjacent' (āśrita) to the 'northern direction,' — which since the Vedas and Pāṇini has signified Greater gandhāra. (Kirfel, W. Das *Purāṇa Pañcalakṣaṇa*.1927.Bonn : K. Schroeder.) This can be construed as a reference to a migration of the sons of Pracetas towards the northern direction to become kings of the mleccha states. The son of Yayati's third son, Druhyu, was Babhru, whose son and grandsons were Setu, Arabdha, Gandhara, Dharma, Dhr̥ta, Durmada and Praceta. It is notable that Pracetas is related to Dharma and Dhr̥ta, who are the principal characters of the Great Epic, the Mahābhārata. It should be noted that a group of people frequently mentioned in the Great Epic are the mleccha, an apparent designation of a group within the country, with Bhāratam janam (Bhārata people). This is substantiated by the fact that Bhagadatta, the king of Pragjyotiṣa is referred to as mleccha and he is also said to have ruled over two yavana kings (2.13).

**Melakkha, island-dwellers, lapidaries**

According to the great epic, Mlecchas lived on islands: "*sa sarvān mleccha nṛpatin sāgara dvīpa vāsinah, aram āhāryàm àsa ratnāni vividhāni ca, andana aguru vastrāṇi maṇi muktam anuttamam, kāñcanam rajatam vajram vidrumam ca mahādhanam*: (Bhima) arranged for all the mleccha kings, who dwell on the ocean islands, to bring varieties of gems, sandalwood, aloe, garments, and incomparable jewels and pearls, gold, silver, diamonds, and extremely valuable coral… great wealth." (MBh. 2.27.25-27). The reference to gems, pearls and corals evokes the semi-precious and precious stones, such as carnelian and agate, of Gujarat traded with Mesopotamian civilization. According to Sumerian records from the Agade Period (Sargon, 2373-2247 BC), Sumerian merchants traded with people from (at least) three named foreign places: Dilmun (now identified as the island of Bahrain in the Persian Gulf); Magan (a port on the coastline between the head of the Persian Gulf and the mouth of the Sindhu river); and Meluhha. Mentions of trade with Meluhha become frequent in Ur III period (2168-2062 BCE) and Larsa dynasty (2062- 1770 BCE). To the end of the Sarasvati Civilization period, the trade declines dramatically attesting to Meluhha being the Sarasvati Civilization. By Ur III Period, Meluhhan workers residing in Sumeria had Sumerian names, leading to a comment: '…three hundred years after the earliest textually documented contact between Meluhha and Mesopotamia, the references to a distinctly foreign commercial people have been replaced by an ethnic component of Ur III society' This is an economic presence of Meluhhan traders maintaining their own village for a considerable span of time.(Parpola, Simo, Asko Parpola, and Robert H. Brunswig, Jr., 1977, "TheMeluhha Village — Evidence of Acculturation of Harappan Traders in Late Third Millenium Mesopotamia?", *Journal of the Economic and Social History of the Orient*, Volume 20, Part II.)

The epic also refers to the pāṇḍava Sahadeva's conquest of several islands in the sea with mleccha inhabitants.

A reference also to the salty marshes of Rann of Kutch in Gujarat (and also, perhaps, the Makran coast, south of Karachi), may also be surmised, where settlements and fortifications such as Amri Nal, Allahdino, Dholavira (Kotda) Sur-kota-da, and Kanmer have been excavated – close to the Sarasvati River Basin as the River traversed towards the Arabian ocean. *Kathāsaritsāgara* (tr. CH Tawney, 1880, Calcutta; rep. New Delhi, 1991), I, p. 151 associates mleccha with Sind. Mleccha kings paid tributes of sandalwood, aloe, cloth, gems, pearls, blankets, gold, silver and valuable corals.

Nakula conquered western parts of Bhāratavarṣa teeming with mleccha (MBh.V.49.26: *yah pratīcīm diśam cakre vaśe mlecchagaṇāyutām sa tatra nakulo yoddhā citrayodhī vyavasthitah*). Bṛhatsamhitā XIV.21 refers to lawless mleccha who inhabited the west: *nirmaryādā mlecchā ye paścimadiksthit āsteca*. A Buddhist chronicle, *āryaManjuśrī Mūlakalpa* [ed. Ganapati Śāstri, II, p. 274] associates pratyanta (contiguous)with mlecchadeśa in western Bhāratavarṣa: *paścimām diśīm āsṛtya rājāno mriyate tadā ye 'pi pratyantavāsinyo mlecchataskarajīvinah.* (trans. 'Then (under a certain astrological combination) the kings who go to the west die; also inhabitants of pratyanta live like the mlecchas and taskara.')

This metaphor defines the region fit for yajna. This metaphor also explains the movements of mleccha, such as kamboja-yavana, pārada-pallava along the Indian Ocean Rim as sea-faring merchants from Meluhha. This parallels the hindu-bauddha continuum exemplified by the Mathura lion capital with śrivatsa and Angkor Wat (Nagara vātika) as the largest Viṣṇu mandiram in the world, together with celebration of Bauddham in many parts of central, eastern and southeastern Asian continent. Mleccha were at no stage described in any text as people belonging to one ethnic, religious or linguistic group. This self-imposed restriction evidenced by all writers of the early Indian cultural tradition – Veda, Bauddha, Jaina alike – is of fundamental significance in understanding that mleccha constituted the core of the people on the banks of Rivers Sarasvati and Sindhu and were the principal architects, artisans, workers, and people, in general, of the Sarasvati-Sindhu Civilization throughout its stages of evolution through phases in modes of production – pastoral, agricultural, industrial – and interactions with neighbors, trading in surplus food products and artefacts generated and sharing cultural attributes/ characteristics.

Various terms are used to describe mleccha social groups and communities: *pratyantadeśa* (*Arthaśāstra* VII.10.16), *paccantimā janapada* (*Vinaya Piṭaka* V.13.12, vol. I, p. 197), *aṭavi, aṭavika* (DC Sircar, *Selected Inscriptions*, vol. I, 'Thirteenth Rock Edict Shābhāzgaṛhī, text line 7, p.37; 'Khoh Copper Plate Inscription of Saimkshobha', text line 8; *Arthaśāstra* VII.10.16; VII.4.43: *mlecchaṭavi* who were considered a threat to the state; *Arthaśāstra* IX.2.18-20 mentions *aṭavibala*, troops from forests as one of six types of troops at the disposal of a ruler). Some mleccha lived in border areas and forests, e.g. *pratyanta nṛpatibhir* (frontier kings: JF Fleet, CII, vol. II, 'Allahabad Posthumous Pillar Inscription of Samudragupta, text line 22, p. 116) cf. Arthaśāstra– a 4[th] century BCE text — I.12.21; VII.14.27; XIV.1.2; *mleccha jāti* are: *bheda, kirāta, śabara, pulinda*: *Amarakośa* II.10.20, a fifth century CE text).

In many Persian inscriptions Yauna, Gandhāra and Saka occur together. [For e.g., DC Sircar, *Selected Inscriptions*, no.2 'Persepolis Inscription on Dārayavahuṣ (Darius c. 522-486 BCE),' lines 12-13, 18, p.7; no. 5, 'Perseplis Inscription of Khshayārshā (Xerxes c. 486-465)', lines 23, 25-6, p. 12].

Thus, *yavana* may be a reference to people settled in the northwest Bhāratavarṣa (India).

There are references to Mleccha (that is, śaka, Yavana, Kamboja, Pahlava) in Bāla Kāṇḍa of the *Valmiki Rāmāyaṇa* (1.54.21-23; 1.55.2-3). *Taih asit samvrita bhūmih śakaih-Yavana miśritaih || 1.54-21 || taih taih Yavana-Kamboja barbarah ca akulii kritaah || 1-54-23 || tasya humkaarato jātah Kamboja ravi sannibhah | udhasah tu atha sanjatah Pahlavah śastra panayah || 1-55-2|| yoni deśāt ca Yavanah śakri deśāt śakah tathā | roma kupeṣ u Mlecchah ca Haritah sa Kiratakah || 1-55-3 ||.Kāmboja Yavanān caiva śakān paṭṭaṇāni ca | Anvīkṣya Varadān caiva Himavantam vicinvatha || 12 || — (Rāmāyaṇa 4.43.12)*

The Yavanas here refer to the Bactrian Yavanas (in western Oxus country), and the Sakas here refer to the Sakas of Sogdiana/Jaxartes and beyond. The Vardas are the same as Paradas (*Hindu Polity*, 1978, p 124, Dr K. P. Jayswal; *Goegraphical Data in Early Purana*, 1972, p 165, 55 fn, Dr M. R.Singh). The Paradas were located on river Sailoda in Sinkiang (MBh II.51.12; II.52.13; VI.87.7 etc) and probably as far as upper reaches of river Oxus and Jaxartes (Op cit, p 159-60, Dr M. R.Singh).

Vanaparva of Mahābhārata notes: "…...Mlechha (barbaric) kings of the śaka-s, Yavanas, Kambojas, Bahlikas etc shall rule the earth (i.e India) un-rightously in Kaliyuga…" *viparīte tadā loke purvarūpān kṣayasya tat || 34 || bahavo mechchha r\ājānah pṛthivyām manujādhipa | mithyanuśāsinah pāpa mṛṣavadaparāṇah || 35 || āndrah śakah Pulindaśca Yavanaśca narādhipāh | Kamboja Bahlikah śudrastathābhīra narottama || 36||* MBH 3/188/34-36). Anushasanaparava of Mahābhārata affirms that Mathura, was under the joint military control of the Yavanas and the Kambojas (12/102/5). *Tathā Yavana Kambojā Mathurām abhitaś ca ye ete niyuddhakuśalā dākshiinātyāsicarminah*. Mahābhārata speaks of the Yavanas, Kambojas, Darunas etc as the fierce mleccha from Uttarapatha : *uttaraścāpare mlechchha jana bharatasattama. || 63 || Yavanashcha sa Kamboja Daruna mlechchha jatayah. | — (MBH 6.11.63-64) They are referred to as papakritah (sinful): uttara pathajanmanah kirtayishyami tanapi. | Yauna Kamboja Gandharah Kirata barbaraih saha. || 43 || ete pāpakṛtāstatra caranti pṛrthivīmimām. | śvakakabalagridhraṇān sadharmaṇo narādhipa. || 44 || —* (MBh 12/207/43-44) http://en.wikipedia.org/wiki/Invasion_of_India_by_Scythian_Tribes#Establishment_of_Mlechcha_Kingdoms_in_Northern_India

Yavana are descendants of Turvaśu, one of the four sons of Yayāti. The sons were to rule over people such as Yavana, Bhoja and Yādava (MBh. 1.80.23-4; Matsya Purāṇa 34.29-30). Yavana, descendants of Turvaśu are noted as meat-eaters, sinful and hence, anārya. [MBh. trans. PC Roy, vol. I, p. 179] These people were brought over the sea safely by Indra (RV 6.20.12). In the Mahābhārata, sons of Anu are noted as mleccha. ṛgveda notes that Yadu and Turvaśa are dāsa (RV 10.62.10):

*sanema te vasā navya indra pra pūrava stavanta enā yajnaih*

*sapta yat purah śarma śāradīr dadruiśa dhan dāsīh purukutsāya śikṣan*

*tvam vrdha indraprvyarja bhūr varivasyann uśane kāvyāya*

*parā navavāstvam anudeyam mahe pitre dadātha svam napātam*

*tvam dhunir indra dhunimtṛṇor āpah sīrā na sravantīh*

*pra yat samudram ati śūra parśi pāraya turvaśam yadum svasti*

RV 6.020.10 (Favoured) by your proection, Indra, we solicit new (wealth); by this adoration men glorify you at sacrifices, for that you have shattered with your bolt the seven cities of śarat, killing the opponents (of sacred rites), killing the opponents (of sacred rites), and giving (their spoils) to Purukutsa. [Men: puravah = manuṣyah; śarat = name of an asura].

RV 6.020.11 Desirous of opulence, you, Indra, have been an ancient benefactor of Us'anas, the son of Kavi; having slain Navavāstva, you have given back his own grandson, who was (fit) to be restored o the grandfather.

RV 6.020.12 You, Indra, who make (your enemies) tremble, have caused the waters, detained by Dhuni, to flow like rushing rivers; so, hero, when, having crossed the ocean, you have reached the shore, you have brought over in safety Turvas'a and Yadu. [*samudram atipraparṣi* = samudram atikramya pratirṇo bhavasi = when you are crossed, having traversed the ocean, you have brought across Turvaśa and Yadu, both standing on the future shore, *samudrapāretiṣṭhantau apārayah*].

Nandana, another commentator of *Mānava Dharma śāstra*. X.45, defines *āryavāc as samskṛtavāc*. Thus, according to Medhātithi, neither habitation nor mleccha speech is the ground for regarding groups as Dasyus, but it is because of their particular names Barbara etc., that they are so regarded. These people were brought over the sea safely by Indra, as noted by this ṛca. This ṛca also notes that Yadu and Turvaśa (are) dāsa; and that Turvaśu is a son of Yayāti. The sons of Yayāti were to rule over people such as Yavana, Bhoja and Yādava. Turvaśu and Yadu crossed the oceans to come into Bhāratavarṣa. In this ṛca., 'samudra' can be interpreted only as an ocean. The ocean crossed by Indra, may be not too far from Sindhu. Sindhu is a 'natural ocean frontier' in ṛgveda. Given the activities of the Meluhha along the Makran Coast (300 km. south of Mehergarh, in the neighbourhood of Karachi), Gulf of Kutch and Gulf of Khambat, (evidence? *Turbinella pyrum* —śankha-bangle found in a woman's grave in Mehergarh, dated to c. 6500 BCE, yes 7[th] millennium BCE; the type of shell found nowhere else in the world excepting the coastline of Sindhu sāgara upto to the Gulf of Mannar).

The ocean referred to may be the ocean in the Gulf of Kutch and was situated with a number of dvīpas. In places north of Lamgham district, i.e. north bank of river Kabul, near Peshawar were regions known as Mi-li-ku, the frontier of the mleccha lands. [S. Beal, 1973, *The Life of Hiuen Tsiang*, New Delhi, p 57; cf. NL Dey, *Geographical Dictionary of India*, p. 113 for an identification of Lamgham (Lampakā) 20 miles north-west of Jalalabad.] *Harivamśa* 85.18-19 locates the mleccha in the Himalayan region and mleccha are listed with yavana, śaka, darada, pārada, tuṣāra, khaśa and pahlava in north and north-west Bhāratavarṣa: *sa viv ṛddho yad ā rāj ā yavan ānām mah ābalāh tata enam nṛpā mlecch āh sams'rity ānuyayaus tad ā Śakās tuṣār ā daradāh pāradās tan:gaṇāh khasśāh pahlavāh śataśaścānye mlecch ā haimavat ās tathā. Matsya Purāṇa* 144.51-58 provides a list. Pracetā had a hundred sons all of whom ruled in mleccha regions in the north. [Matsya Purāṇa 148.8-9; Bhāgavata Purāṇa IX.23.16.] Bhīṣma Parvan of Mahābhārata notes that mleccha jāti people lived in Yavana, Kāmboa, Dāruṇā regions and are listed together with several other peoples of the northern and north-western parts of Bhāratavarṣa (MBh. VI.10.63-66: *uttarāścāpare mlecchā janā bharatasattama yavanāśca śaka, kāmbojā dārun.ā mlecchajātayah*). In *Rāmāyaṇa* IV.42.10, Sugrīva is asked to search for Sītā in the northern lands of mleccha, pulinda, sūrasena, praṣalā, bhārata, kuru, madraka, kamboja and yavana before proceeding to Himavat: *tatra mlecchān pulindāmśūrasen āmś tathaiva ca prasthalān bharatāmścaiva kurūmsśca saha madraih*. Mlecchas came from the valley adjoining the Himalaya. [Rājatarangiṇī, VII. 2762-64.]

   When Sagara, son of Bāhu, was prevented from destroying śaka, Yavana, Kāmboa, Pārada and Pāhlava after he recovered his kingdom, Vasiṣṭha, the family priest of Sagara, absolved these people of their duties but Sagara commanded the Yavana to shave the upper half of their heads, the Pārada to wear long hair and Pahlava to let their beards grow. Sagara also absolved them of their duty to offer yajna to agni and to study the Veda. [Vāyu Purāṇa 88.122. 136- 43; Brahmāṇḍa Purāṇa 3.48.43-49; 63.119-34.] This is how these Yavana, Pārada and Pahlava also became mleccha. [Viṣṇu Purāṇa 4.3.38-41.] The implication is that prior to Sagara's command, these kṣatriya communities did respect Vasiṣṭha as their priest, studied the Veda and performed yajna. [Harivamśa 10.41-45.] Śaka who were designated as kings of mleccha jāti by Bhaṭṭa Utpala (10[th] century) in his commentary on Bṛhatsamhitā, were defeated by Candragupta II. That the mleccha were also adored as ṛṣi is clear from the verse of *Bṛhatsamhitā* 2.15: *mlecchā hi yavanās teṣu samyak śāstram kadam sthitam ṛṣivat te 'pi pūjyante kim punar daivavid dvijāh* (The yavana are mleccha, among them this science is duly established; therefore, even they (although mleccha) are honoured as ṛṣi; how much more (praise is due to an) astrologer who is a brāhmaṇa'). *Bṛhatsamhitā* 14.21 confirms that the yavana, śaka and pahlava lived on the west. Similarly, Konow notes that Sai-wang (Saka King) mentioned in Chinese accounts should be interpreted as Saka Muruṇda and the territory he occupied as Kāpiśa. [Sten Konow, *CII*, vol. II, pp. xx ff; Sten Konow, EI, no. 20 'Taxila Inscription of the Year 136', vol. XIV, pp. 291-2.] Śaka migrated to Bhāratavarṣa through Arachosia via the Bolan Pass into the lower Sindhu, a region called Indo_Scythia by Greek geographers and called śaka-dvīpa in Bhāratiya texts. [EJ Rapson, ed., 1922, *Cambridge History of India*, vol. I, Ancient India, Cambridge, p. 564.] Another view expressed by Thomas is that the migration was through Sindh and the valley of the Sindhu River. [FW Thomas, 'Sakastana', JRAS, 1906, p. 216.] Kalhaṇa notes that Jalauka, a son of Aśoka took possession of Kāśmīra, advanced as far as Kanauj, after crushing a horse of mleccha. [Rājataraṅgiṇī, 1.107-8.] Greek invasions occurred later, during the reign of Puṣyamitra Śunga (c. 185-150 BCE). The regions inhabited by the 'milakkha' could be the Vindhyan region. The term, 'mleccha' of which 'milakkha' is a variant, could as well have denoted the indigenous people (Nahali?) or of Bhāratavarṣa who had lived on the Sarasvati River basin and who moved towards other parts of Bhāratavarṣa after the gradual desiccation of the river, over a millennium, between c. 2500 and 1500 BCE. Medhātithi, commenting on the verse of Manu, defines a language as mleccha : *asad avidyam ān\arthās ādhu śabdatayā vāk mleccha ucyate yathā śabarāṇām kirātānām anyeyām va antyānām*: Medhātithi on *Mānava Dharmaśāstra* X.45 – 'Language is called mleccha because it consists of words that have no meaning or have the wrong meaning or are wrong in form. To this class belong the languages of such low-born tribes as the Śabara-s, Kirāta and so forth…'… He further proceeds to explain that āryavāc is refined speech and the language of the inhabitants of āryāvarta, but only of those who belong to the four varṇa-s. The others are called Dasyus.: ibid. – *āryavāca āryāvartam vāsinas te cāturvarṇy ādanyajātīyatvena prasiddhas tadā dasyava ucyante* 'Arya (refined) language is the language of the inhabitants of āryāvarta. Those persons being other than the four varṇa-s are called Dasyus.'

In Dhammapada's commentary on Petuvathu, Dwaraka is associated with Kamboja as its Capital or its important city.[ The Buddhist Concepts of Spirits, p 81, Dr B. C. Law.] See evidence below:

"*Yasa asthaya gachham Kambojam dhanharika/ ayam kamdado yakkho iyam yakham nayamasai// iyam yakkham gahetvan sadhuken pasham ya/ yanam aaropyatvaan khippam gaccham Davarkān iti* "
[Buddhist Text *Khudak Nikaya* (P.T.S)]

Mleccha who came to the Rājasūya also included those from forest and frontier areas (MBh. III. 48.19: *sāgarān ūpagāmścaiva ye ca paṭṭaṇavāsinah simhalān barbarān mlecchān ye ca jān:galavāsinah*). Bhīmasena proceeded east towards Lohitya (Brahmaputra) and had conquered several mleccha people who bestowed on him wealth of various kinds (MBh. II.27.23-24: *suhmānāmādhipam caiva ye ca sāgaravāsinah sarvān mlecchagaṇāmścaiva vijigye bharatarṣabhah evam bahu vidhān deśān vijitya pavanātmajah vasu tebhya upādya lauhityam agad balī*. [NL Dey, *Geographical Dictionary*, p. 115.]

Celebrations at the Kalinga capital of Duryodhana were attended by preceptors and mleccha kings from the south and east of Bhārata (MBh. XII.4.8: *ete cānye ca bahavo dakṣināṃ diśām āśritah mlecchā āryāśca rājānah prācyodicyāśca bhārata*).

Bhāgadatta, the great warrior of Prāgjyotiṣa accompanied by mleccha people inhabiting marshy regions of the sea- coast (*sāgarānūpavāsibhih*), attends the Rājasūya of Yudhiṣṭhira (MBh. II.31.9-10: *prāgjyotiṣaśca nṛpatir bhagadatto mahāyaśāh saha sarvais tathā mlecchaih sāgarānūpavāsibhih*). This is perhaps a reference ot the marshy coastline of Bengal. *Amarakośa* II, Bhūmivarga – 6: pratyanto mlecchadeśah syāt; Sarvānanda in his commentary, ṭīkāsarvasva, elaborates that mleccha deśa denotes regions without proper conduct such as Kāmarūpa: *bhāratavarṣasyāntadeśah śiṣṭācārā rahitah kāmarūpādih mlecchadeśāh* [Nāmalingānuśāsana, with commentary ṭīkāsarvasva, of Sarvānanda (ed. Ganapati śāstri)]; he also cites Manu that where four varṇa-s are not established that region is mlecchadeśa. A contemporary of Harṣavardhana was Bhāskaravarman of Kāmarūpa; this king was supplanted by another dynasty founded by śālastambha who was known as a mleccha overlord. [SK Chatterji, 1950, Kirāta-jana-kṛti --The Indo-Mongoloids: Their contributions to the and culture of India, Journal of Royal Asiatic Society of Bengal, Vol. XVI, pp.143-253.]

Meluhha, Mleccha areas: Sarasvati River Basin and Coastal Regions of Gujarat, Baluchistan

Meluhha referred to in Sumerian and old Akkadian texts refers to an area in Sarasvati Civilization; Asko and Simo Parpola add: '…probably, including NW India with Gujarat as well as eastern Baluchistan'.[ WF Leemans, Foreign Trade in the Old Babylonian Period, 1960; 'Trade Relations on Babylonia', Journal of Economic and Social History of the Orient, vol. III, 1960, p.30 ff. 'Old Babylonian Letters and Economic History', Journal of Economic and Social History of the Orient, vol. XI, 1968, pp. 215-26; J. Hansam, 'A Periplus of Magan and Meluhha', Bulletin of the School of Oriental and African Studies, vol. 36, pt. III, 1973, pp. 554-83. Asko and Simo Parpola, 'On the Relationship of the Sumerian Toponym Meluhha and Sanskrit Mleccha', *Studia Orientalia*,vol. 46, 1975, pp. 205-38.]

Imports from Meluhha into Mesopotamia included the following commodities which were found in north-western and western Bhāratavarṣa: copper, silver, gold, carnelian, ivory, uśu wood (ebony), and another wood which is translated as 'sea wood' – perhaps mangrove wood on the coasts of Sind ad Baluchistan. [J. Hansman, 'A Periplus of Magan and Meluhha', Bulletin of the School of Oriental and African Studies, vol. 36, pt. III, 1973, pp. 560.] The Ur texts specifically refer to 'seafaring country of Meluhha'' and hence, Leemans' thesis that Meluhha was the west coast (modern state of Gujarat) of Bhārata. The Lothal dockyard had fallen into disuse by c.1800 BCE, a date when the trade between Mesopotamia and Meluhha also ended. [WF Leemans, 'Old Babylonian Letters and Economic History', Journal of Economic and Social History of the Orient, vol. XI, 1968, pp. 215-26. P. Aalto, 1971, 'Marginal Notes on the Meluhha Problem,' Professor KA Nilakanta Sastri Felicitation Volume, Madras, pp. 222-23.] In Leemans' view, Gujarat was the last bulwark of the (Indus or Sarasvati) Civilization. Records refer to Meluhhan ships docking at Sumer. There were Meluhhans in various Sumerian cities; there was also a Meluhhan town or district at one city. The Sumerian records indicate a large volume of trade; according to a Sumerian tablet, one shipment from Meluhha contained 5,900 kg of copper (13,000 lbs, or 6 ½ tons)! The bulk of this trade was done through Dilmun, not directly with Meluhha. In our view, the formative stages of the Civilization also had their locus in the coastal areas – in particular, the Gulf of Khambat, Gulf of Kutch and Makran coast, as evidenced by the wide shell-bangle, dated to c. 6500 BCE, made of turbinella pyrum or śankha, found in Mehergarh, 300 miles north of the Makran coast.

**Tanana mleccha**

A Jaina text, Avasyaka Churani notes that ivory trade was managed by mleccha, who also traveled from Uttaravaha to Dakshinapatha.[ Jain, 1984, Life in Ancient India as Described in the Jain Canon and Commentaries (6th century BC - 17th century AD, p. 150.] Guttila Jataka (ca.4th cent.) makes reference to itinerant ivory workers/traders journeying from Varanasi to Ujjain. [Cowell, 1973, Jatakas Book II, p. 172 ff.] The phrase, *tanana mleccha* may be related to: (i) tah'nai, 'engraver' mleccha; or (ii) tana, 'of (mleccha) lineage'. 1. See Kuwi. Tah'nai 'to engrave' in DEDR and Bsh. Then, thon, 'small axe' in CDIAL: DEDR 3146 *Go.* (Tr.) tarcana , (Mu.) tarc- to scrape; (Ma.) tarsk- id., plane; (D.) task-, (Mu.) tarsk-/tarisk- to level, scrape (*Voc.*1670).

**Sea-faring merchants/artisans of Meluhha**

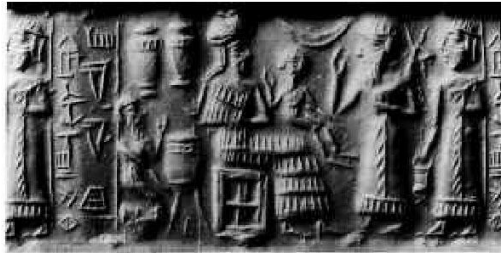

Akkadian. Cylinder seal Impression. Inscription records that it belongs to 'S'u-ilis'u, Meluhha interpreter', i.e., translator of the Meluhhan language (EME.BAL.ME.LUH.HA.KI) The Meluhhan being introduced carries an goat on his arm. Musee du Louvre. Ao 22 310, Collection De Clercq 3rd millennium BCE. The Meluhhan is accompanied by a lady carrying a kamaṇḍalu.

Since he needed an interpreter, it is reasonably inferred that Meluhhan did not speak Akkadian.

Antelope carried by the Meluhhan is a hieroglyph: mlekh 'goat' (Br.); mṛeka (Te.); mēṭam (Ta.); meṣam (Skt.) Thus, the goat conveys the message that the carrier is a Meluhha speaker. A phonetic determinant.mrṛeka, mlekh 'goat'; Rebus: melukkha Br. mēlh 'goat'. Te. mṛeka (DEDR 5087)  meluh.h.a

"While Prof. Thomson maintained that a Munda influence has probably been at play in fixing the principle regulating the inflexion of nouns in Indo-Aryan vernaculars, such influence appeared to be unimportant to Prof. Sten Konow… Prof. Przyluski in his papers, translated here, have tried to explain a certain number of words of the Sanskrit vocabulary as fairly ancient loans from the Austro-Asiatic family of languages. He has in this opened up a new line of enquiry. Prof. Jules Bloch in his article on Sanskrit and Dravidian, also translated in this volume, has the position of those who stand exclusively for Dravidian influence and has proved that the question of the Munda substratum in Indo-Aryan cannot be overlooked…In 1923, Prof. Levi, in a fundamental article on *Pre-Aryen et Pre-Dravidian dans Vinde* tried to show that some geographical names of ancient India like Kosala-Tosala, Anga-Vanga, Kalinga-Trilinga, Utkala-Mekala and Pulinda-Kulinda, ethnic names which go by pairs, can be explained by the morphological system of the Austro-Asiatic languages. Names like Accha-Vaccha, Takkola-Kakkola belong to the same category. He concluded his long study with the following observation, " We must know whether the legends, the religion and the philosophical thought of India do not owe anything to this past. India has been too exclusively  examined from the Indo-European standpoint. It ought to be remembered that India is a great maritime country… the movement which carried the Indian colonization towards the Far East… was far from inaugurating a new route…Adventurers, traffickers and missionaries profited by the technical progress of navigation and followed under better conditions of comfort and efficiency, the way traced from time immemorial, by the mariners of another race, whom Aryan or Aryanised India despised as savages."
In 1926, Przyluski tried to explain the name of an ancient people of the Punjab, the Udumbara, in a similar way and affiliate it to the Austro-Asiatic group. (cf. *Journal Asiatique*, 1926, 1, pp. 1-25, Un ancien peuple du Pendjab — les Udumbaras: only a portion of this article containing linguistic discussions has been translated in the Appendix of this book.) In another article, the same scholar discussed some names of Indian towns in the geography of Ptolemy and tried to explain them by Austro-Asiatic forms…Dr. J. H. Hutton, in an interesting lecture on the Stone Age Cult of Assam delivered in the Indian Museum at Calcutta in 1928, while dealing with some prehistoric monoliths of Dimapur, near Manipur, says that " the method of erection of these monoliths is very important, as it throws some light on the erection of prehistoric monoliths in other parts of the world. Assam and Madagascar are the only remaining parts of the world where the practice of erecting rough stones still continues….The origin of this stone cult is uncertain, but it appears that it is to be mainly imputed to the Mon-Khmer intrusion from the east In his opinion the erection of these monoliths takes the form of the lingam and yoni. He thinks that the Tantrik form of worship, so prevalent in Assam, is probably due to " the incorporation into Hinduism of a fertility cult which preceded it as .the religion of the country. The dolmens possibly suggest distribution from South India, but if so, the probable course was across the Bay of Bengal and then back again westward from further Asia. Possibly the origin was from Indonesia whence apparently the use of supari (areca nut) spread to India as well as the Pacific." (From the Introduction by PC Bagchi and SK Chatterjee, 1 May 1929).

Kuiper notes: " …a very considerable amount (say some 40%) of the New Indo-Aryan vocabulary is borrowed from Munda, either via Sanskrit (and Prākṛt), or via Prākṛt alone, or directly from Munda; wide-branched and seemingly native, word-families of South Dravidian are of Proto-Munda origin; in Vedic and later Sanskrit, the words adopted have often been Aryanized, resp. Sanskritized. "In view of the intensive interrelations between Dravidian, Munda and Aryan dating from pre-Vedic times even individual etymological questions will often have to be approached from a Pan-Indic point of view if their study is to be fruitful. It is hoped that this work may be helpful to arrive at this all-embracing view of the Indian languages, which is the final goal of these studies." F.B.J. Kuiper, 1948, *Proto-Munda Words in Sanskrit*, Amsterdam, Verhandeling der Koninklijke Nederlandsche Akademie Van Wetenschappen,
Afd. Letterkunde, Nieuwe Reeks Deel Li, No. 3, 1948, p.9
http://www.scribd.com/doc/12238039/mundalexemesinSanskrit

Emeneau notes: "In fact, promising as it has seemed to assume Dravidian membership for the Harappa language, it is not the only possibility. Professor W. Norman Brown has pointed out (The United States and India and Pakistan, 131-132, Cambridge, Harvard University Press, 1953) that Northwest India, i.e. the Indus Valley and adjoining parts of India, has during most of its history had Near Eastern elements in its political and cultural make-up at least as prominently as it had true Indian elements of the Gangetic and Southern types. [M.B.Emeneau, India as a Linguistic Area [Lang. 32, 1956, 3-16; LICS, 196, 642-51; repr. In Collected papers: Dravidian Linguistics Ethnology and Folktales, Annamalai Nagar, Annamalai University, 1967, pp. 171-186.] The passage is so important that it is quoted in full: 'More ominous yet was another consideration. Partition now would reproduce an ancient, recurring, and sinister incompatibility between Northwest and the rest of the subcontinent, which, but for a few brief periods of uneasy cohabitation, had kept them politically apart or hostile and had rendered the subcontinent defensively weak. When an intrusive people came through the passes and established itself there, it was at first spiritually closer to the relatives it had left behind than to any group already in India. Not until it had been separated from those relatives for a fairly long period and had succeeded in pushing eastward would I loosen the external ties. In period after period this seems to have been true. In the third millennium B.C. the Harappa culture in the Indus Valley was partly similar to contemporary western Asian civilizations and partly to later historic Indian culture of the Ganges Valley. In the latter part of the next millennium the earliest Aryans, living in the Punjab and composing the hymns of the Rig Veda, were apparently more like their linguistic and religious kinsmen, the Iranians, than like their eastern Indian contemporaries. In the middle of the next millennium the Persian Achaemenians for two centuries held the Northwest as satrapies. After Alexander had invaded India (327/6-325 B.C.) and Hellenism had arise, the Northwest too was Hellenized, and once more was partly Indian and partly western. And after Islam entered India, the Northwest again was associated with Persia, Bokhara, Central Asia, rather than with India, and considered itself Islamic first and Indian second. The periods during which the Punjab has been culturally assimilated to the rest of northern India are ew if any at all. Periods of political assimilation are almost as few; perhaps a part of the fourth and third centuries B.C. under the Mauryas; possibly a brief period under the Indo-Greek king menander in the second century B.C.; another brief period under the Muslim kingdom of Delhi in the last quarter of the twelfth century A.D.; a long one under the great Mughals in the sixteenth and seventeenth centuries A.D.; a century under the British, 1849-1947.'

"Though this refers to cultural and political factors, it is a warning that we must not leap to linguistic conclusions hastily. The early, but probably centuries-long condition in which Sanskrit, a close ally of languages of Iran, was restricted to the northwest (though it was not the only language there) and the rest of India was not Sanskritic in speech, may well have been mirrored earlier by a period when some other language invader from the Near East-a relative of Sumerian or of Elamitic or what not-was spoken and written in the Indus Valley-perhaps that of invaders and conquerors-while the indigenous population spoke another language-perhaps one of the Dravidian stock, or perhaps one of the Munda stock, which is now represented only by a handful of languages in the backwoods of Central India.

"On leaving this highly speculative question, we can move on to an examination of the Sanskrit records, and we find in them linguistic evidence of contacts between the Sanskrit-speaking invaders and the other linguistic groups within India…the early days of Indo-European scholarship were without benefit of the spectacular archaeological discoveries that were later to be made in the Mediterranean area, Mesopotamia and the Indus Valley… This assumption (that IE languages were urbanized bearers of a high civilization) led in the long run to another block-the methodological tendency of the end of the nineteenth and the beginning of the twentieth century to attempt to find Indo-European etymologies for the greatest possible portion of the vocabularies of the Indo-European languages, even though the object could only be achieved by flights of phonological and semantic fancy… very few scholars attempted to identify borrowings from Dravidian into Sanskrit…The Sanskrit etymological dictionary of Uhlenbrck (1898-1899) and the Indo-European etymological dictionary of Walde and Pokorny (1930-1932) completely ignore the work of Gundert (1869), Kittel (1872, 1894), and Caldwell (1856,1875)… It is clear that not all of Burrow's suggested borrowings will stand the test even of his own principles…'India' and 'Indian' will be used in what follows for the subcontinent, ignoring the political division into the Republic of India and Pakistan, and, when necessary, including Ceylong also… the northern boundary of Dravidian is and has been for a long time retreating south before the expansion of Indo-Aryan… We know in fact from the study of the non-Indo-European element in the Sanskrit lexicon that at the time of the earliest Sanskrit records, the R.gveda, when Sanskrit speakers were localized no further east than the Panjab, there were already a few Dravidian words current in Sanskrit. This involves a localization of Dravidian speech in this area no lather than three millennia ago. It also of course means much bilingualism and gradual abandonment of Dravidian speech in favor of IndoAryan over a long period and a great area-a process for which we have only the most llsd of evidence in detail. Similar relationships must have existed between Indo-Aryan and Munda and between Dravidian and Munda, but it is still almost impossible to be sure of either of these in detail… The Dravidian languages all have many Indo-Aryan items, borrowed at all periods from Sanskrit, Middle Indo-Aryan and Modern Indo-Aryan. The Munda languages likewise have much Indo-Aryan material, chiefly, so far as we know now, borrowed rom Modern Indo-Aryan, thogh this of course llsdes items that are Sanskrit in form, since Modern Indo-Aryan borrows from Sanskrit very considerably. That Indo-Aryan has borrowed from Dravidian has also become clear. T. Burrow, The Sanskrit Language, 379-88 (1955), gives a sampling and a statement of the chronology involved. It is noteworthy that this influence was spent by the end of the pre-Christian era, a precious indication for the linguistic history of North India: Dravidian speech must have practically ceased to exist in the Ganges valley by this period… Most of the languages of India, of no matter which major family, have a set of retroflex, cerebral, or domal consonants in contrast with dentals. The retroflexes include stops and nasal certainly, also in some languages sibilants, lateral, tremulant, and even others. Indo-Aryan, Dravidian, Munda and even the far northern Burushaski, form a practically solid bloc characterized by this phonological feature… Even our earliest Sanskrit records already show phonemes of this class, which are, on the whole, unknown elsewhere in the Indo-European field, and which are certainly not Proto-Indo-European. In Sanskrit many of the occurrences of retroflexes are conditioned; others are explained historically as reflexes of certain Indo-European consonants and consonant clusters. But, in fact, in Dravidian it is a matter of the utmost certainty that retroflexes in contrast with dentals are Proto-Dravidian in origin, not the result of conditioning circumstances… it is clear already that echo-words are a pan-Indic trait and that Indo-Aryan probably received it from non-Indo-Aryan (for it is not Indo-European)… The use of classifiers can be added to those other linguistic traits previously discussed, which establish India as one linguistic area ('an area which includes languages belonging to more than one family but showing traits in common which are found not to belong to the other members of (at least) one of the families') for historical study. The evidence is at least as clear-cut as in any part of the world… Some of the features presented here are, it seems to me, as 'profound' as we could wish to find… Certainly the end result of the borrowings is that the languages of the two families, Indo-Aryan and Dravidian, seem in many respects more akin to one another than Indo-Aryan does to the other Indo-European languages. (We must not, however, neglect Bloch's final remark and his reasons therefor: *'Ainsi donc, si profondes qu'aient ete les influences locales, lls n'ont pas conduit l'aryen de l;inde… a se differencier fortement des autres langues indo-europeennes.'*)" M.B.Emeneau, Linguistic Prehistory of India PAPS98 (1954). 282-92; Tamil Culture 5 (1956). 30-55; repr. In Collected papers: Dravidian Linguistics Ethnology and Folktales, Annamalai Nagar, Annamalai University, 1967, pp. 155-171.

The profundity of these observations by Emeneau and Bloch will be tested through clusters of lexemes of an *Indian Lexicon*, which relate to the archaeological finds of the civilization.

Tamil and all other Dravidian languages have been influenced by Sanskrit language and literature. Swaminatha Iyer [Swaminatha Iyer, 1975, Dravidian Theories, Madras, Madras Law Journal Office] posits a genetic relationship between Tamil and Sanskrit. He cites GU Pope to aver that several Indo-European languages are linguistically farther away from Sanskrit than Dravidian. He cites examples of Tamil and Sanskrit forms of some glosses: hair: mayir, s'mas'ru; mouth: vāya, vā c; ear: śevi, śrava; hear: kēḻ keṇ (Tulu), karṇa; walk: śel, car; mother: āyi, yāy (Paiśāci). Evaluating this work, Edwin Bryant and Laurie Patton note: "It is still more simple and sound to assume that the words which need a date of contact of the fourth millennium BCE on linguistic grounds as loan words in Dravidian might be words originally inherited in Dravidian from the Proto-speech which was the common ancestor of both Dravidian and Indo-Aryan…It will be simpler to explain the situation if both Indo-Aryan and Dravidian are traced to a common language family. In vocables they show significant agreement. In phonology and morphology the linguistic structures agree significantly. It requires a thorough comparative study of the two language families to conduct a fuller study. " Bryant, Edwin and Laurie L. Patton, 2005, The Indo-Aryan controversy: evidence and inference in Indian history, Routledge, p.197.

The influence of Vedic culture is profoundly evidenced in early sangam texts. **K. V. Sarma**, 1983, "Spread of Vedic Culture in Ancient. South India" in The **Adyar** Library **Bulletin**, **1983**, 43:1.

**Proto-Munda continuity and Language X**

Sources of OIA agricultural vocabulary based on Masica (1979)

|     | Percentage |
| --- | --- |
| • IE/Iir | 40% |
| • Drav | 13% |
| • Munda | 11% |
| • Other | 2% |
| • Unknown | 34% |
| • Total | 100% |

Hence, a Language X is postulated; Language 'X' to explain a large number of agriculture-related words with no IE cognates: Colin Masica, 1991, Indo-Aryan Languages, Cambridge Univ. Press

Since there is cultural continuity in India from the days of Sarasvati civilization, it is possible to reconstruct Language X by identifying isoglosses in the linguistic area.

Contributions of the following language/archaeology scholars have followed up on these insights of Sylvan Levi, Jules Bloch and Jean Przyluski published over 90 years ago: Emeneau, MB, Kuiper, FBJ, Masica, CP, Southworth F. [Emeneau, MB, 1956, India as a linguistic area, in: *Language*, 32.3-16; Kuiper, FBJ, 1967, The genesis of a linguistic area, *Indo-Iranian Journal* 10: 81-102; Masica, Colin P., 1976, *Defining a linguistic area, South Asia*, Chicago, University of Chicago Press; Franklin Southworth, 2005, *Linguistic Archaeology of South Asia*, Routledge Curzon]

Resemblances between two or more languages (whether typological or in vocabulary) can be due to genetic relation (descent from a common ancestor language), or due to borrowing at some time in the past between languages that were not necessarily genetically related. When little or no direct documentation of ancestor languages is available, determining whether a similarity is genetic or areal can be difficult.

**Further researches**

In addition to studies in the evolution of and historical contacts among Indian languages, further researches are also needed in an archaeological context. Karl Menninger cites a remarkable instance. In the Indian tradition, finger signals were used to settle the price for a trade transaction. Finger gestures were a numeric cipher!

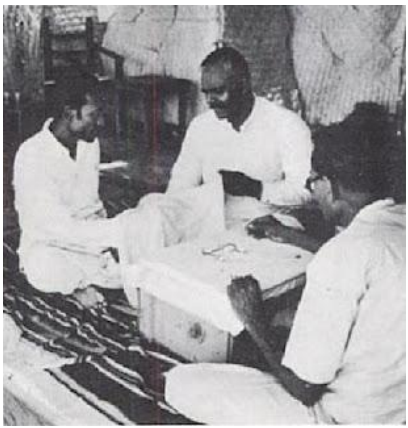

A pearl merchant of South India settling price for a pearl using finger gestures under a handkerchief. Cited in Karl Menninger, 1969, *Number words and number symbols: a cultural history of numbers*, MIT Press, p.212. http://tinyurl.com/26ze95s

Further work on the nature of the contacts between Indian artisans and their trade associates, say, in Meluhhan settlements in the Persian Gulf region, may unravel the the nature of long-distance contacts. Could it be that the Indus language and writing were Indus Artisans' cryptographic messaging system for specifications of artifacts made in and exported from Meluhha?

**Linguistics and archaeo-metallurgy: Identifying meluhha words and matching hieroglyphs with lexemes of archaeo-metallurgy**

Indian Hieroglyphs are identified. This announcement in Archaeometallurgy may be taken as Kitty Hawk flight demo or Jean-Francois Champollion demonstration of Egyptian hieroglyphs. Announcing that Indus script, an unsolved puzzle for over 150 years since the first discovery of a seal by the archaeologist of British India, Alexander Cunningham, are composed of Indian hieroglyphs, the book is said to detail in about 800 pages what could possibly be the earliest invention of writing.

Hundreds of Indian hieroglyphs have been identified in the context of the bronze age and the rebus readings are comparable to the rebus method employed for Egyptian hieroglyphs. The book has related the invention of writing to the invention of bronze-age technologies of mixing copper with other ores such as arsenic, zinc, tin to create alloys like bronze, brass, pewter. The book relates the hieroglyphs to the lexemes of Indian *sprachbund*.

Use of iron was also attested during the bronze age.[9] A surprising find in matching meluhha lexemes with hieroglyphs is that, as noted by the late Gregory Possehl, an Indus archaeologist, iron was also used, though archaeo-metallurgy evidence for iron-smelters have not so far been discovered in th civilization area.

Archaeo-metallurgy studies of Sarasvati (Indus) Civilization have made some progress[10]. These studies have to be elaborated further to identify the processes of continuity evidenced by the iron smelters identified in the Ganga valley. D.K. Chakraborti and James Muhly argue that metallurgy of tin was well developed in Indus (Sarasvati) Civilization. The use of zinc as evidenced by the svastika glyphs is surprising and has to be explained further in archaeo-metallurgy context. One possibility is that zinc-bearing ores were used to create bronze alloy ingots and tools/vessels.

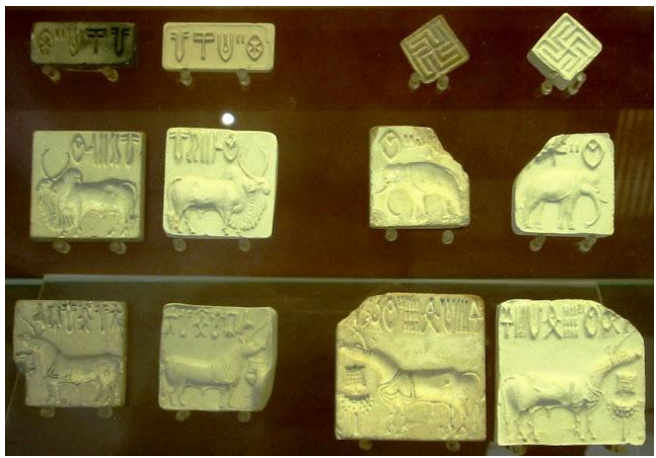

Mainstream linguistics has no way to determine a range of dates for this *sprachbund* (language union). I submit that the language union relates to the bronze age inventions and trade which is complemented by and necessitated the invention of writing. In my view, the script records the archaeometallurgy transactions using lexes of Indian *sprachbund*. The tradition continues in ancient Indian mints which produced the early punch-marked coins. The tradition is also evidenced on the Rampurva copper bolt hieroglyphs, Sohgaura copper plate inscription and Sanchi s'rivatsa hieroglyph.

http://upload.wikimedia.org/wikipedia/commons/thumb/3/32/IndusValleySeals.JPG/220px-IndusValleySeals.JPG

Indian hieroglyphs typical of the 3rd millennium BCE: Elephant:ibh rebus:ib 'iron'; koḍ 'one-horned heifer' rebus: koḍ 'smithy';sathiya 'svastika glyph' rebus: satiya,jasta 'zinc'; adar 'zebu' rebus: aduru 'unsmelted metal or ore'; pattar 'trough' rebus: 'smiths' guild'; kaṇḍ karṇaka 'rim of jar' rebus: kaṇḍ karṇaka 'furnace account scribe', ayakara 'fish+crocodile' rebus: 'metal-smith' etc.

Hundreds of such examples are discussed in *Indian hieroglyphs*[11] demonstrating that Indian hieroglyphs constitute a writing system for meluhha language and are rebus representations of archaeo-metallurgy lexemes.

After scholars review this work which covers hieroglyphs used in about 5000 indus script inscriptions of the corpora and validate the rebus readings, a mile-stone would have been recorded in the study of ancient civilizations. The identification of Indian hieroglyphs may, then, turn out to be as historic as the decoding of Egyptian hieroglyphs by Jean-Francois Champollion and be the foundation for further studies in (a) the evolution of languages of the Indian *sprachbund* (language union) and (b) archaeo-metallurgical traditions.

**Indus script corpora and business transactions of jangad, 'entrustment note'**

A function is posited for specific seals of Indus script corpora (with young bull + lathe hieroglyphs) that the hieroglyphs used on such seals were intended to connote 'entrustment notes' (जांगड़ jāngāḍ) for trade transactions from Meluhha and constituted an improvement in documentation and control of guild (corporation) transactions over the earlier system of tokens, tallies and bullae. The military guard who delivered products into the treasury is called jangaḍiyo (Gujarati). The business tradition of jangad continues even today among diamond merchants/cutters of India. The monograph is organized in the following sections:

• Young bull + lathe hieroglyphs on Indus seals
• Indus writing system in Susa and harosheth hagoyim, 'smithy of nations'

The following note 'Seal m0296 read rebus' provides a remarkable reinforcement of the reading of the hieroglyph sangaḍa 'lathe/portable furnace'. The lexemes of Western Pahadi and Pashto with the semantics 'chain' provide this phonetic reinforcement: śã́gal, śã́gaḍ 'chain' (WPah.) زغرہ zˊg̠haraˊh, s.f. (3rd) Chain armour. Pl. ئ ev. زغ یالَ zˊg̠har vālaey, s.m. (1st) A man in armour. Pl. ئ ī. (Pashto) சங்கிலி¹ caṅkili , n. < śṛṅkhalaā. [M. caṅ- kala.] 1. Chain, link; தொடர். சங்கிலிபோ லீர்ப்புண்டு (சேதுபு. அகத். 12). 2. Land-measuring chain, Gunter's chain 22 yards long; அளவுச் சங்கிலி. (C. G.) 3. A superficial measure of dry land=3.64 acres; ஓர் நிலவளவு. (G. Tn. D. I, 239). 4. A chain-ornament of gold, inset with diamonds; வயிரச்சங்கிலி என்னும் அணி. சங்கிலி நுண்டொடர் (சிலப். 6, 99). 5. Hand-cuffs, fetters; விலங்கு. ശൃംഖല šr̥ṅkhala S. A chain, Tdbh. சஙஅல 341.(Malayalam) சங்காட்டம் caṅkāṭṭam , n. < saṅ-ghaṭṭa. Union, intercourse; சேர்க்கை. சங்காட்டந்தவிர்த்து (தேவா. 655, 1). http://www.docstoc.com/docs/118797742/sangad
Seal m0296 read rebus

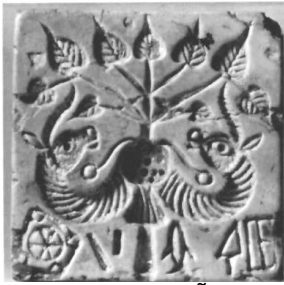 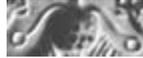 śā́gal, śā́gaḍ 'chain' (WPah.) زغره *zʿgharaʿh*, s.f. (3rd) Chain armour. Pl. ي‎ *ey*. زغر يالَي‎ *zʿghar yālaey*, s.m. (1st) A man in armour. Pl. ي‎ *ī*. (Pashto) sankhalā (f.) [cp. Sk. śṛṅkhalā] a chain Th 2, 509. aṭṭhi° a chain of bones, skeleton A iii.97. (Pali) śṛṅkhala 12580 śṛṅkhala m.n. ' chain ' MārkP., °lā -- f. VarBṛS., śṛṅkhalaka -- m. ' chain ' MW., ' chained camel ' Pāṇ. [Similar ending in mékhalā -- ] Pa. saṅkhalā -- , °likā -- f. ' chain '; Pk. saṁkala -- m.n., °lā -- , °lī -- , °liā -- , saṁkhalā -- , siṁkh°, siṁkalā -- f. ' chain ', siṁkhala -- n. ' anklet '; Sh. šăṅāli̯ f., (Lor.) š*lṅāli, šiṅ° ' chain ' (lw .with š -- < śṛ -- ), K. hŏkal f.; S. saṅgharu m. ' bell round animal's neck ', °ra f. ' chain, necklace ', saṅghāra f. ' chain, string of beads ', saṅghirī f. ' necklace with double row of beads '; L. saṅglī f. ' flock of bustard ', awāṇ. saṅgul ' chain '; P. saṅgal m. ' chain ', ludh. suṅgal m.; WPah.bhal. śaṅgul m. ' chain with which a soothsayer strikes himself ', śaṅgli f. ' chain ', śiṅkhal f. ' railing round a cow -- stall ', (Joshi) śā́gaḷ ' door -- chain ', jaun. śā́gal, śā́gaḍ 'chain'; Ku. sãglo ' doorchain ', gng. śā́ṇaw ' chain '; N. sā̃lo ' chain ', °li ' small do. ', A. xikali, OB. siṅkala, B. sikal, sikli, chikal, chikli, (Chittagong) hĭol ODBL 454, Or. sāṅk(h)uḷā, °li, sāṅkoḷi, sikaḷã̄, °ḷi, sikuḷā, °ḷi; Bi. sīkaṛ ' chains for pulling harrow ', Mth. sĩ̄kaṛ; Bhoj. sĩ̄kar, sīkarī ' chain ', OH. sāṁkaḍa, sīkaḍa m., H. sā̃kal, sā̃kar, °krī, saṅkal, °klī, sikal, sīkar, °krī f.; OG. sāṁkalu n., G. sā̃kaḷ, °kḷī f. ' chain ', sā̃kḷū n. ' wristlet '; M. sā̃k(h)aḷ, sāk(h)aḷ, sā̃k(h)ḷī f. ' chain ', Ko. sāṁkaḷ; Si. säkilla, hä°, ä° (st. °ili -- ) ' elephant chain '. śṛṅkhalayati. WPah.kṭg. (kc.) śáṅgəḷ f. (obl. -- i) ' chain ', J. śā́gaḷ f., Garh. sā̃gaḷ. (CDIAL 12580).

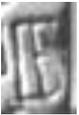 The last sign on epigraph 5477 and 1554 (m296 seal) is read as: kole.l = smithy, temple in Kota village (Ko.)

Glyph: 'piece': guḍá—1. — In sense 'fruit, kernel' cert. ← Drav., cf. Tam. koṭṭai 'nut, kernel'; A. goṭ 'a fruit, whole piece', °ṭā 'globular, solid', guṭi 'small ball, seed, kernel'; B. goṭā 'seed, bean, whole'; Or. goṭā 'whole, undivided', goṭi 'small ball, cocoon', goṭāli 'small round piece of chalk'; Bi. goṭā 'seed'; Mth. goṭa 'numerative particle' (CDIAL 4271) Rebus: koṭe 'forging (metal)(Mu.) Rebus: goṭī f. 'lump of silver' (G.) goṭi = silver (G.) koḍ 'workshop' (G.). Glyph: 'two links in a chain': Vikalpa: kaḍī a chain; a hook; a link (G.); kaḍum a bracelet, a ring (G.) Rebus: kaḍiyo [Hem. Des. kaḍaio = Skt. sthapati a mason] a bricklayer; a mason; kaḍiyaṇa, kaḍiyeṇa a woman of the bricklayer caste; a wife of a bricklayer (G.) The stone-cutter is also a mason.

Glyptic elements of m296 seal impression: 1. Two heads of one-horned heifers; 2. ligatured to a pair of rings and a standard device; 3. ligatured to a precise count of nine leaves. Read rebus: koḍiyum 'heifer, rings on neck'; rebus: koḍ 'workshop' (Kuwi.G.); dula 'pair' (Kashmiri); rebus: dul 'cast metal' (Mu.) lo, no 'nine' (B.); loa 'ficus religiosa' (Santali); rebus: loh 'metal' (Skt.); loa 'copper' (Santali) sangaḍa 'jointed animals' (Marathi); sangaḍa 'lathe' (G.) Part of the pictorial motif is thus decoded rebus: loh dul koḍ 'metal cast(ing) smithy turner (lathe) workshop '. Part of the inscription is read rebus: *ayaskāṇḍa kole.l* 'smithy, excellent quantity of iron'.

The stem in the orthographic composition relates to *sangaḍa* 'lathe/furnace' (yielding crucible stone ore nodules), the standard device which is depicted frequently in front of 'one-horned heifer'. Rebus: *sangāta* 'association, guild' or, *sangatarāsu* 'stone-cutter' (Telugu). The 'globules' glyphic joining the two ringed necks of a pair of one-horned heifers may connote: goṭi. It may connote a forge.

kamaḍha = *ficus religiosa* (Skt.); kamar.kom 'ficus' (Santali) rebus: kamaṭa = portable furnace for melting precious metals (Te.); kampaṭṭam = mint (Ta.) Vikalpa: Fig leaf 'loa'; rebus: loh '(copper) metal'. loha-kāra 'metalsmith' (Skt.).

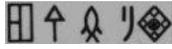Text on m296 seal.
Glyphs: ayas 'fish'. Rebus: aya 'metal'. Glyph: kaṇḍa 'arrow' Rebus: 'stone (ore)metal'; kaṇḍa 'fire-altar'. ayaskāṇḍa is explained in Panini as 'excellent quantity of iron'. It can also be explained as 'metal of stone (ore) iron.'

Thus, the three text sign sequence can be explained rebus as smithy for metal of stone (ore) iron.

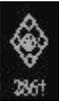 taṭṭai 'mechanism made of split bamboo for scaring away parrots from grain fields (Ta.); taṭṭe 'a thick bamboo or an areca-palm stem, split in two' (Ka.) (DEDR 3042) toṭxin, toṭ.xn goldsmith (To.); taṭṭāṉ 'gold- or silver-smith' (Ta.); taṭṭaravāḍu 'gold- or silver-smith' (Te.); *ṭhaṭṭakāra 'brass-worker' (Skt.)(CDIAL 5493). Thus, the glyph is decoded: taṭṭara 'worker in gold, brass'.

This is a complex, ligatured glyph with a number of glyphic elements. May denote a cast metal (copper) worker guild working with 4 types of pure metal and alloyed ingots (copper + arsenic/tin/zinc).

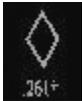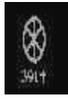

Glyphic element: erako nave; era = knave of wheel. Glyphic element: āra 'spokes'. Rebus: āra 'brass' as in ārakūṭa (Skt.) Rebus: Tu. eraka molten, cast (as metal); eraguni to melt (DEDR 866) erka = ekke (Tbh. of arka) aka (Tbh. of arka) copper (metal); crystal (Ka.lex.) cf. eruvai = copper (Ta.lex.) eraka, er-aka = any metal infusion (Ka.Tu.); erako molten cast (Tu.lex.) Glyphic element: kund opening in the nave or hub of a wheel to admit the axle (Santali) Rebus: kundam, kund a sacrificial fire-pit (Skt.) kunda 'turner' kundār turner (A.); kũdār, kũdāri (B.); kundāru (Or.); kundau to turn on a lathe, to carve, to chase; kundau dhiri = a hewn stone; kundau murhut = a graven image (Santali) kunda a turner's lathe (Skt.)(CDIAL 3295)

Glyphic element: 'corner': *khuṇṭa2 ' corner '. 2. *kuṇṭa -- 2. [Cf. *khōñca -- ] 1. Phal. khun ' corner '; H. khū̃ṭ m. ' corner, direction ' (→ P. khūṭ f. ' corner, side '); G. khū̃ṭrī f. ' angle '. <-> X kōṇa -- : G. khuṇ f., khū̃ṇo m. ' corner '.2. S. kuṇḍa f. ' corner '; P. kūṭ f. ' corner, side ' (← H.). (CDIAL 3898). Rebus: khū̃ṭ 'community, guild' (Mu.)

ചങ്ങാതം čaṇṇāḏam (Tdbh.; സംഘാതം) 1. Convoy, guard; responsible Nāyar guide through foreign territories. ച. പോരുക to accompany as such. ച. പോന്ന വാരിയര്, എന്നെ ച'വും കൂട്ടി അയച്ചു TR. 2. income of Rājas from granting such guides; grant of land to persons liable to such service ച. കൊടുക്ക. 3. companion പന്നിയും കാട്ടിയും ച'മായി CG.—met. കംസനെ കൊന്ന ഗോപാലനെ കംസനു ച'മാ ക്കുവാന് CG. to send him along, to kill likewise. ചങ്ങാതി (C. Te. സ —) companion, തുണക്കാ രന്; friend വീണാല് ചിരിക്കാത്ത ച. ഇല്ല, ച. നന്നെങ്കില് കണ്ണാടി വേണ്ട prov. ച. യായുള്ളു പണ്ടുപണ്ടേ CC.—also fem. ച ങ്ങാതിമാരായുള്ള അംഗനമാര് CG.; vu. എ) ന്റെ ചങ്ങായിച്ചീ TP. (Voc.) See also: ചങ്ങു V1. a small chain to which to hang keys etc. ചങ്ങാടം čaṇṇāḏam (Tu. ജംഗാല, Port. Jangada). Ferryboat, junction of 2 boats. ച. കെ ട്ടുക; ച'ത്തില് കേറി TR. തോണികള് ച' ങ്ങള് വഞ്ചികള് പടവുകള് Bhr. also rafts. (Malayalam) sangaḍa 'lathe/portable furnace'; rebus: जाकड़ ja:kaṛ (nm) on approval (purchases); —का माल goods/articles on approval. (H.lexicon) sangara [fr. saṇ+gr̥1 to sing, proclaim, cp. gāyati & gīta] 1. a promise, agreement J iv.105, 111, 473; v.25, 479 (Pali) angadia 'courier' (Gujarati)cf. jangaḍia 'military guard accompanying treasure into the treasury' (Gujarati) Ta. aṅkāṭi bazaar, bazaar street. Ma. aṅṅāṭi shop, bazaar. Ko. aṅga·ḍy id. To. ogoḍy bazaar (? < Badaga). Ka. aṅgaḍi shop, stall. Koḍ. aṅgaḍi id. Tu. aṅgaḍi id. Te. aṅgaḍi id. Kol. aṅgaḍi bazaar. Nk. aṅgāṛi id. Nk. (Ch.) aṅgāṛ market. Pa. aṅgoḍ courtyard, compound. / ? Cf. Skt. aṅgaṇa- courtyard.(DEDR 35). cf. semantics of 'tying up, packaging': जखडणें [ jakhaḍaṇēṃ ] v c To tighten or draw tight. 2 To tie up or to: (as a beast to a stake.) It is in both senses generally used with another verb, as बांधणें, टाकणें, धरणें, राखणें.जखडबंदी [jakhaḍabandī] f (जखडणें & P) Tying up (as a beast to a stake). v कर g. of o.: also tied up state. Also fig. rigidly binding, obliging, confining: also bound state. 2 unc. Tying and binding; wrapping and fastening; packing up. (Marathi) Semantics of bailiff 'custody, charge, moving': జంగమము [ jiṅgamamu ] jangamamu. [Skt.] adj. Moveable, not stationary. తిరుగునది. జంగమ కట్టుబడె a temporary bailiff. జంగమనగము, (Vasu. iii. 249.) జంగమరావము, or జంగమాద్రి rolling rock, a moving hill. P. i. 202; iii. 62. n. A moveable or chattel; property, personalty. Cattle, cows, sheep, &c. జంగముడు jangamuḍu. n. Jangam, or worshipper of Basava. L. XIV. 210. జంగమత్వము jangamatvamu. n. Moveableness, locomotion. G. ix. 121. (Telugu) jaṅgama ' moving ' AitUp. [√gam] Pa. jaṅgama -- , Pk. jaṁgama -- ; Si. däṅguma ' motion, going to and fro '.(CDIAL 5079)Cognate gloss is Pali sanghāta or sanghāta is variously interpreted but, generally, with reference to the semantics of 'accumulation, aggregation': Sangharaṇa (nt.) [=saṇharaṇa] accumulation J iii.319 (dhana°).Sangharati [=saṇharati] 1. to bring together, collect, accumulate J iii.261; iv.36 (dhanaṃ), 371; v.383. <-> 2. to crush, to pound J i.493.Sanghāta [fr. saṇ+ghāṭeti, lit. "binding together"; on etym. see Kern, Toev. ii.68] 1. a raft J ii.20, 332 (nāvā°); iii.362 (id.), 371. Miln 376. dāru° (=nāvā°) J v.194, 195. -- 2. junction, union VvA 233. -- 3. collection, aggregate J iv.15 (upāhana°); Th 1, 519 (papañca°). Freq. as aṭṭhi° (cp. sankhalā etc.) a string of bones, i. e. a skeleton Th 1, 570; DhA iii.112; J v.256. -- 4. a weft, tangle, mass (almost="robe," i. e. sanghāṭī), in taṇhā° -- paṭimukka M i.271; vāda° -- paṭimukka M i.383 (Neumann "defeat"); diṭṭhi° -- paṭimukka Miln 390. <-> 5. a post, in piṭṭha° door -- post, lintel Vin ii.120.Sanghāta [saṇ+ghāta] 1. striking, killing, murder Vin i.137; D i.141; ii.354; M i.78; A ii.42 sq. -- 2. knocking together (cp. sanghaṭṭeti), snapping of the fingers (acchara°) A i.34, 38; J vi.64. -- 3. accumulation, aggregate, multitude PvA 206 (aṭṭhi° mass of bones, for the usual °sanghāṭa); Nett 28. -- 4. N. of one of the 8 principle purgatories J v.266, 270.Sanghātanika (adj.) [fr. sanghāta or sanghāṭa] holding or binding together M i.322 (+agga -- sangāhika); A iii.10 (id.); Vin i.70 ("the decisive moment" Vin. Texts i.190). (Pali) "The second translator (of Ārya Sanghāta Sūtra) into Chinese rendered the title of the sutra in Chinese as The Sutra of the Great Gathering of the Holy Dharma. (In Chinese, Ta chi hui cheng fa ching in the Wade-Giles transliteration system, or Ta ji-hui zheng-fa jing in Pinyin.)"http://en.wikipedia.org/wiki/Sanghata_Sutra cf. saṁgraha m. ' collection ' Mn., ' holding together ' MBh. [√grah]Pa. saṅgaha -- m. ' collection ', Pk. saṁgaha -- m.; Bi. sãgah 'building materials'; Mth. sãgah 'the plough and all its appurtenances', Bhoj. har -- sãga; H. sãgahā 'collection of materials (e.g. for building)'; <-> Si. saṅgaha ' compilation ' ← Pa.*saṁgrahati 'collects' see sáṁgr̥hṇāti.(CDIAL 12852). S. saṅgu m. 'body of pilgrims' (whence sãgo m. 'caravan'), L. P. saṅg m.(CDIAL 12854).

Allograph:
sanghaṭṭana (nt.)bracelet (?) SnA 96 (on Sn 48). angada [cp. Sk. angada; prob. anga + da that which is given to the limbs] a bracelet J v.9, 410 (citt°, adj. with manifold bracelets). (Pali)aṅgada n. 'bracelet on upper arm' R. [← Muṇḍa Kuiper PMWS 124] Pa. aṅgada -- n., Pk. aṁgaya -- n., Si. aňguva.(CDIAL 117)A. śā̃k (phonet. x -- ) 'bracelet made of shells' AFD 187.(CDIAL 12263). அங்கதம்² aṅkatam n. < aṅgada. Bracelet worn on the upper arm; வாகுவலயம். புயவரை மிசை . . .அங்கதம் (திருவிளை. மாணிக். 12).

Allograph? சங்கடம்² caṅkaṭam , n. < Port. jangada. Ferry-boat of two canoes with a platform thereon; இரட்டைத்தோணி. (J.)jangada id. (Portuguese)

खोंड [khōṇḍa] m A young bull, a bullcalf. (Marathi)kurī´ 'colt, calf'(CDIAL 3245). కోడియ [kōḍiya] Same as కోడె. kōḍe. [Tel.] n. A bullcalf. కోడెదూడ. A young bull. కాడిమరపదగినదూడ. Plumpness, prime. తరుణము. కోడుకోడెయలు a pair of bullocks. కోడె adj. Young. కోడెనాగు a young snake, one in its prime. "కోడెనాగమును బలుగుల రేదుతన్ని పోవుతెరంగు" రామా. vi. కోడెకాడు kōḍe-kāḍu. n. A young man. పడుచువాడు. A lover విటుడు.Te. kōḍiya, kōḍe young bull; adj. male (e.g. kōḍe dūḍa bull calf), young, youthful; kōḍekāḍu a young man. Kol. (Haig) kōḍē bull. Nk. khoṛe male calf. Koṇḍa kōḍi cow; kōṛe young bullock. Pe. kōḍi cow. Manḍ. kūḍi id. Kui kōḍi id., ox. Kuwi (F.) kōḍi cow; (S.) kajja kōḍi bull; (Su. P.) kōḍi cow.(DEDR 2129). Rebus: A. kundār, B. kūdār, °ri, Or. kundāru; H. kŭderā m ' one who works a lathe, one who scrapes ', °rī f., kŭdernā ' to scrape, plane, round on a lathe '.kundakara m. ' turner ' W. [Cf. *cundakāra -- : kunda -- 1, kará -- 1](CDIAL 3297) कोंदणपट्टी [ kōndaṇapaṭṭī ] f The strip of beaten or drawn gold used in setting gems.कोंदण [ kōndaṇa ] n (कोंदणें) Setting or infixing of gems. 2 Beaten or drawn gold used in the operation. 3 The socket of a gem.(Marathi) కుందనము [ kundanamu ] kundanamu. [Tel.] n. Solid gold, fine gold. అపరంజి. kunda1 m. ' a turner's lathe ' lex. [Cf. *cunda -- 1]N. kŭdnu ' to shape smoothly, smoothe, carve, hew ', kŭduwā ' smoothly shaped '; A. kund ' lathe ', kundiba ' to turn and smooth in a lathe ', kundowā ' smoothed and rounded '; B. kŭd ' lathe ', kŭdā, kōdā ' to turn in a lathe '; Or. kū̃nda ' lathe ', kūdibā, kū̃d° ' to turn ' (→ Drav. Kur. kū̃d ' lathe '); Bi. kund ' brassfounder's lathe '; H. kunnā ' to shape on a lathe ', kuniyā m. ' turner ', kunwā m.(CDIAL 3295). Allographs: Konta 'a pennant, standard' (cp. kunta) J vi.454; DA i.244; SnA 317.(Pali)Sk. kunta lance? a. கோடு-. 1. [K.kōḍu]Crookedness, flexure, obliquity; வளைவு. 2. Partiality, bias; நடுநிலை நீங்குகை. கோடிறீக் கூற்றம் (நாலடி, 5). 3. [K. kōḍu.] Tusk; யானை பன்றிகளின் தந்தம். மத்த யானையின் கோடும் (தேவா. 39, 1). 4. Horn; விலங்கின் கொம்பு. கோட்டிடை யாடினை கூத்து (திவ். இயற். திருவிருத். 21). b. கோடு[K. kōḍu, M. kōṭu.] Summit of a hill, peak; மலைச்சிகரம். பொற்கோட் டிமயமுரும் (புறநா. 2, 24). 15. Mountain; மலை. குமரிக் கோடும் (சிலப். 11, 20). கோடர் kōṭar , n. < கோடு². Peak, summit of a tower; சிகரம். கோடரி நீண்மதிற் கோட் டாறு (இறை. 23, உதா. 199). c. கோடு[K. kōḍu.] Branch of a tree; மரக்கொம்பு. (பிங்.) 8. Body of a lute; யாழ்த்தண்டு. மகர யாழின் வான்கோடு தழீஇ (மணி. 4, 56). கோடரம்¹ kōṭaram , n. prob. id. 1. Branch of a tree; மரக்கொம்பு. (பிங்.) Rebus: கோடு[M. kōṭṭa.] Stronghold, fortified place; அரணிருக்கை. (W.)கோட்டம்² kōṭṭam , n. < kōṣṭha. 1.

Room, enclosure; அறை. சுடுமண்ணோங்கிய நெடு நிலைக் கோட்டமும் (மணி. 6, 59). 2. Temple; கோட்டம்; kōṭṭam , n. < gō-ṣṭha. 1. Cow- shed Read on...http://www.docstoc.com/docs/118578044/Indus-script-corpora-and-business-transactions-of-jangad-%E2%80%98entrustment-note%E2%80%99-(S-Kalyanaraman-2012)

**Young bull + lathe hieroglyphs on Indus seals**

A seal impression was found at Tell Umma. This showed the hieroglyphs of 'young bull + lathe', a hieroglyphic set which is common in the Indus script corpora of now over 6000 inscriptions. What did these two hieroglyphs mean? An attempt is made to decode the hieroglyphs reading them rebus in Meluhha (Mleccha) language of the Indian *sprachbund*.

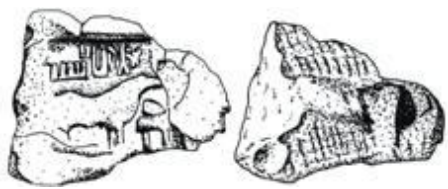

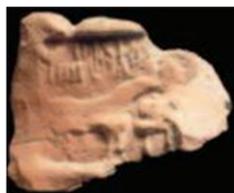

Seal impression of Tell Umma with Indus writing

Impression of a 'unicorn' seal thought to come from Tell Umma. Cited in Gregory L. Possehl, The Middle Asian Interaction Sphere, *Expedition*, UPenn, p.41. Umma (modern Tell Jokha/Djoha) was a Sumerian city state in entral southern Mesopotamia. One-horned heifer. Scheil 1925. Indicative of the receipt of goods from the Sarasvati-Sindhu and of the possible presence of Indus traders in Mesopotamia. Tell Asmar seals, together with ceramics, knobbed ware, etched beads and kidney shaped inlay of bone provide supporting evidence for this possibility. http://www.penn.museum/documents/publications/expedition/PDFs/49-1/Research%20Notes.pdf

See: S. Kalyanaraman, 2011, Decoding Indus script Susa cylinder seal: Susa-Indus interaction areas. http://www.docstoc.com/docs/102138513/Decoding-Indus-Scipt-Susa-cylinder-seal-Susa-Indus-interaction-areas-(Kalyanaraman-2011)

Hypothesis: Tokens as tallies evolved as seals with 'lathe' hieroglyph: 'entrustment receipts'. Functions of Indus seals in evolution of writing system. [Evidence of seal impressions of Kanmer which could be strung together the way tokens were strung together, as demonstrated by Denise Schmand-Besserat.]

The seals with these hieroglyphs may be jangad 'for approval' process/trade transactions (say, between workers' platforms to warehouse or from warehouse to sales agents).

Since modern use of 'heifer' refers to a young cow, I would like to correct the meaning of koḍiyum (G.) as 'young bull, bull-calf'. The cognate term in Telugu: కోడియ [ kōḍiya ] Same as కోడె [ kōḍe ] kōḍe. [Tel.] n. A bullcalf. కోడూడ. A young bull.खोंड [ khōṇḍa ] m A young bull, a bullcalf.(Marathi) ['*Heifer*' may be derived from Old English *heahfore*; related to Greek *poris* calf, bull.]

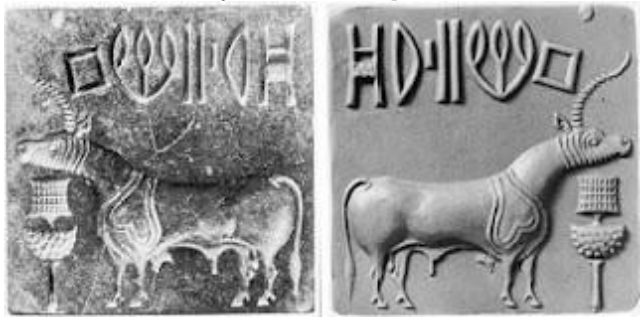

Harappa h006 Seal and impression.

Many seals depict a hieroglyphic composition: (1) one-horned heifer with pannier and neck-rings; and (2) a gimlet/lathe on portable furnace. koḍiyum 'young bull' (G.) koḍ 'horn' (Kuwi) koṭiyum 'rings on neck; a wooden circle put round the neck of an animal' (Gujarati.) खोंडा [khōṇḍā] m A कांबळा of which one end is formed into a cowl or hood (Marathi). kŏdā 'to turn in a lathe'(B.) कोंद kōnda 'engraver, lapidary setting or infixing gems' (Marathi) kūdār 'turner, brass-worker'(Bengali) খোদকার [ khōdakāra ] n an engraver; a carver (Oriya). Glyph: sangaḍa 'lathe' (Marathi) Rebus: जांगड [jāngaḍ] 'a tally of products delivered into the warehouse 'for approval' (Marathi). Rebus: koḍ 'artisan's workshop' (Kuwi) cf. खोट [ khōṭa ] f A mass of metal (unwrought or of old metal melted down); an ingot or wedge.(Marathi)

See: H جاکڑ जाकड़ jākaṛ [fr. S. यतं+कृ; cf. jakaṛnā], s.m. A deposit or pledge left with a vendor for goods brought away for inspection or approval; goods taken from a shop for approval, a deposit or pledge being left; a conditional purchase; articles taken on commission sale;—adv. On inspection, for approval:—jākaṛ-bahī, s.f. Account book of sales subject to approval of goods, &c.:—jākaṛ bećnā, v.t. To sell conditionally, or subject to approval:—jākaṛ le jānā, v.t. To take away goods on inspection, or for approval, leaving a deposit or pledge with the vendor. (Urdu)

Note: The meaning of 'jangad' is well-settled in Indian legal system. Jangad meand "Goods sent on approval or 'on sale or return'… It is well-known that the jangad transactions in this country are very common and often involve property of a considerable value." Bombay High Court
Emperor vs Phirozshah Manekji Gandhi on 13 June, 1934 Equivalent citations: (1934) 36 BOMLR 731, 152 Ind Cas 706 Source: http://www.indiankanoon.org/doc/39008/

Jangad sale is sale on approval and/or consignment basis (that is, taken without definite settlement of purchase).

Discussion of sales on jangad (approval) basis: http://www.lawyersclubindia.com/sc/INDRU-RAMCHAND-BHARVANI-AND-OTHERS-Vs-UNION-OF-INDIA-OTHERS-281.asp

http://indiankanoon.org/doc/1802495/?type=print

[quote]The effect of these terms on the relation between the parties, and the possession of the goods in the hands of the broker, was considered by Madgavkar J. in an unreported judgment in Kanga Jaghirdar & Co. v. Fatehchand Hirachand (1929) O.C.J. Suit No. 1117 of 1928. At that time the relative section of the Indian Contract Act did not contain the expression "mercantile-agent" but only "person". On a consideration of the terms mentioned above the learned Judge came to the conclusion that the possession obtained under a document worded as aforesaid was not juridical possession within the meaning of Section 178 of the Indian Contract Act. As regards the term jangad used in the document the learned Judge observed as follows : "Assuming that jangad in Gujerati ordinarily means 'approval' there is no reason to assume that the goods entrusted jangad are goods to be sold on approval, rather than goods to be shown for approval...The dictionary meaning of the word "jangad" is "approval". As stated by Madgavkar J. in the passage quoted above, having regard to the printed terms in this case, there appears no reason to assume that the diamonds were entrusted to defendants Nos. 1 and 2 to be sold on approval and not that they were given to them to be shown for approval. In my opinion taking the document as a whole, it is clear that they were given to defendants Nos, 1 and 2 to be shown for approval only...It is, therefore, clear that by the delivery of 173 diamonds to him, even on jangad terms, no property can pass to him under Section 24 of the Sale of Goods Act."
[unquote]http://www.indiankanoon.org/doc/1749483/

In one transaction involving diamonds, the case states: "The diamonds were forwarded along with writings titled "ACKNOWLEDGMENT OF ENTRUSTMENT". In the trade they are known as "Jangad" notes. The eight diamonds were forwarded under three identical Jangad notes which also specified the value of the diamonds in Rupees per carat. " http://indiankanoon.org/doc/910302/

Jangad note is typically used in diamond business transactions. "...jangad receipts (letters/bills issued by diamond owners to whom the diamonds are given for the purpose of business prior to sale/export etc." http://www.sitcinfo.com/content/directTaxes/decisions/viewfile.asp?CFN=32591RC.htm

Diamond rough processing: "Each Unit Head sends goods for laser kerfing or sawing. Records of such goods are maintained in registers. Whenever goods are sent to sub contractors for laser operations. Jangads are prepared. Goods meant for laser kerfing are fixed in cassettes and sent to laser division or sub contractors. Diamonds for sawing are sent loose. All goods are sent with details of cut number, quantity, weight, and any other specific instruction that is required."http://www.diamjewels.in/infrastructure.htm

Comment:

It is clear that jangad note is a documentation of a business transaction for property items.

It is remarkable that the trade/pocess transaction tradition is traceable to hieroglyphs of Indus writing. The pronunciation in Gujarati is jangaḍ relatable to jangāḍiyo 'a military guard who accompanies treasure into the treasury'.(Gujarati lexicon) Thus jangaḍ is interpreted as '**acknowledgment of entrustment**' [of property item(s), which are listed by other hieroglyphs on a seal or seal impression.] The word 'angaḍia' comes from jangaḍ and means 'trust'. అంగడి [ aṅgaḍi ] angadi. [Drav.] (Gen. అంగటి Loc. అంగట, plu. అంగళ్లు) n. A shop. అంగడిపెట్టు to open a shop. అంగళ్లవాడ range of shops. అంగట పోకార్చి selling in the shop. అంగడివీధి a market place. ఆ సంగతిని అంగడిలో పెట్టినాడు he revealed or exposed the matter. அங்காடி aṅkāṭi , n. [T.K. aṅgaḍi, M. aṅṅāṭī.] Bazaar, bazaar street; கடை. (சிலப். 14, 179.) Ta. aṅkāṭi bazaar, bazaar street. Ma. aṅṅāṭi shop, bazaar. Ko. aŋga·ḏy id. To. ogoḏy bazaar (? < Badaga). Ka. aṅgaḍi shop, stall. Koḍ. aŋgaḍi id. Tu. aṅgaḍi id. Te. aṅgaḍi id. Kol. aŋgaḍi bazaar. Nk. aŋgāṛi id. Nk. (Ch.) aŋgāṛ market. Pa. aŋgoḍ courtyard, compound. / ? Cf. Skt. aṅgaṇa- courtyard. (DEDR 35). aṅgana n. ' act of walking ' lex., ' courtyard ' R., °aṇa -- n. Kālid. [√aṅg] Pa. aṅgaṇa -- n. ' open space before palace '; Pk. aṁgaṇa -- n. ' courtyard ', K. ãgun dat. -- anas m., S. aṅaṇu m., WPah. bhad. aṅgan pl. -- gnā̃ n., Ku. ānaṇ, N. ānan, B. āṅgan, āṅginā, Or. agaṇā, dial. āṅgan, Bi. ãgan, ãgnā, ẽgnā (BPL 1237), Mth. ãgan, Bhoj. ānan, H. ãgan, ãgnā, agnā m. (X uṭhān s.v. upasthā´na -- ), G. ãgaṇ, ãgṇū n., M. ãgṇẽ n., Ko. āṅgaṇa, °goṇ n., Si. aṅgaṇa, aṅguṇuva. -- Deriv. L. mult. aṅgaṇī f. ' the grains that remain on the threshing floor after division '; G. ãgṇiyũ n. ' open space about a house '.(CDIAL 118) જાંગડ વેચાણ - વહેવારમાં સાચું "એફ" ફોર્મ જરુરી

Source: J.R.Lunagariya, Ahmedabad | Last Updated 12:09[IST](13/12/2010) જાંગડ વેચાણ, "['approval' sale]" is a well-recognized business transaction as note in this Gujarati article. http://business.divyabhaskar.co.in/article/jangad-selling---f-form-need-1644327.html?PRVNX=

That 'jangad' means an "Entrust Receipt" is explained in the rules of Diamond Platform in Mumbai (Bombay): http://www.diamondplatformmumbai.com/CompanyProfilePage.aspx

Semantics of association: sang 'horn', sang 'stone', sang 'association, guild'; sangar 'fortified observation post'.

As words get used in socio-cultural contexts, semantic expansion occurs. It is possible that the alternative or additional meanings were also read rebus when decoding rebus the two hieroglyphs: 'one-horn' and 'portable furnace/lathe'. Some seals show the orthography of a pierced hole glyphs attached to the bottom vessel of the lathe. These could connote stone (ore) with perforation.

The top register of the 'lathe' hieroglyph denotes a gimlet, while the bottom register shows a vessel with smoke emanating : san:ghāḍo, saghaḍī (G.) = firepan; saghaḍī, śaghaḍi = a pot for holding fire (G.) sangaḍ 'lathe/portable furnace'

A word used to denote a horn in some languages of the Indian linguistic area is: saṁga 'horn' śārṅga ' made of horn ' Suśr., n. ' bow ' MBh. [śŕṅga -- ] Pk. saṁga -- ' made of horn '; Paš.lauṛ. ṣāṅg f.(?) ' horn ' (or < śŕṅga -- ). (CDIAL 12409). *śārṅgala ' horned '. [śārṅga -- ] Paš.lauṛ. ṣaṅgala ' a small horn '; K. hǎgul m. ' the stag Cervus wallichii '.(CDIAL 12410). This word saṁga could be a reinforcement of the sang- in: sangaḍ 'lathe'. A rebus word denotes 'stone' : سنگ • (sang) m, Hindi spelling: संग stone, weight; association, union (Persian. Hindi) Hence, the following semantic expansions related to (1) stone (ore) work and (2) stone fortifications (which are characteristic features of many ancient settlement sites of the civilization): Semantics: stone-fortified settlement with enclosures – courtyards -- for trade. Sang, 'stone' (+) angaṇa 'courtyard' cf.angāḍi 'shop'. The word sang may also denote an association, guild. 1. sangatarāsu 'stone-cutter' (Telugu). san:gatarāsū = stone cutter; san:gatarāśi = stone-cutting; san:gsāru karan.u = to stone (S.) 2. Lahnda: sāgaṛh m. ' line of entrenchments, stone walls for defense '.(CDIAL 12845) Sangar connotes a stone fortification or breastwork of stone by defending guards of an army. (Pushto) Sankata 'obstacle' is semantically relatable to the sangar 'defensive observation post'."Sangars - During the Afghan wars of the 'Great Game' tribesmen would hide in the crevices of the rocky mountainsides to observe and to shoot at the British soldiers. These would shoot back, so the positions would be fortified with slabs of rock, embrasures, roofs, camouflage. The Afghan word for these tiny little forts is Sangar. Things have not changed much, and a Sangar is an Observation-Post (OP) which is protected against incoming ordnance and the weather, and from which weapons as well as binoculars could be used. A Sangar is a fortified OP." http://www.defence-structures.com/glossary.htm "Sangar" referred to a stone breastwork, used by the British army on the northwest frontier of India where it was generally impossible to dig protective trenches. 3. सांगडणी [ sāṅgaḍaṇī ] f (Verbal of सांगडणें) Linking or joining together (Marathi). संगति [ saṅgati ] f (S) pop. संगत f Union, junction, connection, association. संगति [ saṅgati ] c (S) pop. संगती c or संगत c A companion, associate, comrade, fellow. संगतीसोबती [ saṅgatīsōbatī ] m (संगती & सोबती) A comprehensive or general term for Companions or associates. संग [ saṅga ] m (S) Union, junction, connection, association, companionship, society. संघट्टणें [ saṅghaṭṭaṇēṁ ] v i (Poetry. संघट्टन) To come into contact or meeting; to meet or encounter. (Marathi) sangāta 'association, guild' M. sǎgaḍṇē ' to link together '. (CDIAL 12855). Pa. kōḍ (pl. kōḍul) horn; Ka. kōḍu horn, tusk, branch of a tree; kōṟ horn Tu. kōḍǔ, kōḍu horn ( (DEDR 2200) கோடு kōṭu Horn; விலங்கின் கொம்பு. கோட்டிடை யாடினை கூத்து (திவ். இயற். திருவிருத். 21). Ko. Kṛ (obl. Kṭ-) horns (one horn is kob), half of hair on each side of parting, side in game, log, section of bamboo used as fuel, line marked out. To. Kwṛ (obl. Kwṭ-) horn, branch, path across stream in thicket. Ka. Kōḍu horn, tusk, branch of a tree; kōṟ horn. Te. Kōḍu rivulet, branch of a river. (DEDR 2200) Standard device often shown in front of a one-horned heifer [read rebus as sāṅgaḍa 'that member of a turner's apparatus by which the piece to be turned is confined and steadied' सांगडीस धरणें To take into linked-ness or close connection with, lit. fig.' (Marathi); rebus: sanghāḍo cutting stone, gilding (Gujarati)] Thus, together, the pair of hieroglyphs may relate to a semantic indication of 1) an engraver working with stone (ore) either for perforated beads or for other metal work converting stone (ore) to metals and alloyed metals and 2) the definition of the place where the work is performed, say, a settlement with stone fortification. Hence, the possible readings of the two glyphs: sāgaṛh koḍ 'artisan-workshop courtyards within stone fortification', i.e. a fortified settlement of lapidaries' guild. Thus, the word sangad may have had two substantive semantics which can be reasonably deduced: 1. Consignment for approval; and 2. Made by turners/engravers/stone (ore) workers' guild, of a fortified (guild) settlement. Further researches are needed on the economic developments in ancient India, following the work, *Economic history of ancient India* (Santosh Kumar Das, 1944). This work presents an evaluation of ancient texts from which business practices can be gleaned. It is necessary to firmly delineate the chronological evolution of production and trade practices of business in the Indian *sprachbund* which had evolved since 3500 BCE within a broad framework of 'trusteeship' evidenced by the practice of 'jangad' or entrustment note, comparable to consignment basis for display of products in a shopfront.

**Chronology of language evolution in Indian *sprachbund***

A falsifiable hypothesis is postulated that it is possible to identify and provide rebus readings from glosses of present-day languagues and can be used to define the contours of Indian *sprachbund* formed from ca. 3500 BCE.

Marathi as we know today is a lot different, yes, from Meluhha of Indian *sprachbund* of 3500 BCE. See Jules Bloch 'La formation de la langue marathe' [The Formation of the Marathi Language], thesis, [1914/1920], Prix Volney. It is part of Indian *sprachbund*. Most languages of India today have existed for millenia. A good account of the ancient history of Marathi vernacular, an apabhraṃśa language of Prākṛtam family is at http://en.wikipedia.org/wiki/Marathi_language
Reconstructing proto-indo-aryan vocabulary will be a good start which will help rebus readings of hieroglyphs on Indus writing. This ain't an article of faith. This can be enlarged as a falsifiable hypothesis. Kuiper's and Colin Masica's work on Munda and Language X are path-breakers.

The challenge for linguists and philologists is to reconstruct that ancient form of mleccha vaacas (meluhha speech). This term for an ancient speech is attested in ancient texts. When a greater challenge of reconstructing Proto-IE has been joined with a lot of * words, there is no reason why a billion people now speaking languages of Indian *sprachbund* cannot join the challenge I have posed.

There are substratum words which are being compiled, e.g. SARVA project of Southworth, UPenn. (and, of course, my Indian Lexicon of 25+ ancient languages). The challenge to all scholars, engaged in studies of ancient people, is to reconstruct that old form which I have hypothesised as Meluhha (mleccha). SARVA project and my lexicon are just a beginning. Just as CDIAL of Turner was a beginning to provide Indo-Aryan vocabulary. A lot of work done subsequently led to the now prevalent *sprachbund* thesis. This has to be carried forward to trace all 'technology' words as technology changes got recorded in *harosheth hagoyim*, 'smithy of nations'.

Many language lexicons with glosses, do retain memories of the past. Some words are not remembered in some dialects, some are in some other dialects, as languages differentiate into dialects and assume the characteristics of a 'language' with unique morphological, phonetic, semantic and grammatical characteristics. This is how many 'substrate' words are identified even in Sumerian for example: words such as sanga 'priest', tibira 'merchant'. The key is to list such substrate words and read them rebus, which is what I have attempted in my Indian Hieroglyphs (2012). This is a work intended to be 'torn apart' -- critically rebutted -- by scholars of various disciplines so that the final hazy picture emerges from the mists of the past. One such attempt is to relate 'trefoil' hieroglyphs of ancient Uruk/Indus and Egypt. When cuneiform had been decoded, there is no reason to be dispirited and gasp about the impossibility of decoding Indus script. It can be and has been decoded in a firm, archaeological context of the bronze age.

The recognition of Indian *sprachbund* itself is a breakthrough, even as it is endorsed by one of the authors who compiled Dravidian Etymological Dictionary. He had to concede that there is a 'Language X'. Now, it has also to be conceded that 'Munda' also existed in 4th millennium 'Iran'. Language X + Munda constitute the crux of the glosses of Meluhha (Mleccha) in so far as they relate to the new inventions of words to define a metallurgical repertoire of the bronze-age.

When a steam-engine was invented, words had to be used to denote the locomotive. A combination of words was used to define the technological innovation coming out of James Watt's discovery of the steaming kettle throwing out the lid: steam + engine.

The history of Indo-aryan languages has NOT yet been fully told. Now the ruling hypothesis is Indian *sprachbund*. Work is ongoing to spell out the contours of this bund. Until Language X and interactions with Munda for Indo-Aryan substratum words are clearly demarcated, the debate will stay joined.

**Indus writing system in Susa and *harosheth hagoyim,*'smithy of nations'**

Susa was a settlement which was founded around 4000 BCE and had yielded a number of tablets inscribed in Proto-Elamite writing with apparent cuneiform script. Based on the evidences of cuneiform records of contacts with Meluhha, Magan and Dilmun, and the context of the evolving bronze-age, it is possible to evaluate Indus writing in Susa and provide a framework for deciphering Indus writing using the underlying Meluhha language. Judges 4:16 reads: "*Now Barak chased the chariots and the army all the way to Harosheth Hagoyim. Sisera's whole army died by the edge of the sword; not even one survived!*" The reason for the use of the phrase harosheth hagoyim 'smithy of nations' is possibly, a widespread presence of smithy in many bronze- and iron-age settlements, some of which might have produced metallic war-chariots. Indus writing which starts ca. 3500 BCE was a sequel to the system of using tokens and tallies to record property transactions. There is evidence for the presence of Meluhhan settlements in Susa and neighboring regions. Susa finds of cylinder seals and seal impressions, bas-relief of spinner and a ritual basin with hieroglyphs of Indus writing can be consistently interpreted in the Meluhhan language in the context of the evolving bronze-age trade ransactions.*kharoṣṭī* (cognate with *harosheth*) was a syllabic writing system with intimations of contacts with Aramaic writing system. Though early evidences of *kharoṣṭī* documents are dated to ca. early 5$^{th}$ century BCE, it is likely that some form of contract documentation using a proto-form of *kharoṣṭī* was perhaps used by artisans and traders, across a vast interaction area which covered a wide geographic area from Kyrgystan (Tocharian) to Haifa (Israel, Seaport on Mediterranean Ocean) – across Sarasvati-Sindu river-basins, Tigris-Euphrates doab, Caspian Sea, and Mediterranean Ocean – of three civilizations Indus, Mesopotamia and Egypt. The evidence of about 6000 Indus script inscriptions provides the details of products traded in this *harosheth hagoyim*, a smithy of nations, indeed.

The argument is presented in the following sections:

Harosheth hagoyim, 'smithy of nations'

Evolving bronze-age and use of tallies, tokens, bullae for archives

Use of tallies, tokens, bullae for archives in the Near East

      Decoding of the identical inscription on the three tablets of Kanmer

Conjecturing a parallel with Sumer bulla envelope system

Presence of Meluhhan merchant (Shu-ilishu cylinder seal)

Evidence for the use of hieroglyphs from Indus writing in Susa

1. Susa pot with '

2. Cylinder seal of Susa with Indus writing

3. Seal impression of Tell Umma with Indus writing

4. Cylinder Seal of Ibni-Sharrum

5. 'Goat-fish' ligatured hieroglyph of Indus writing on Susa vat

6. Bas-relief of spinner with hieroglyphs of Indus writing

7. Susa stamp seals from the Persian Gulf

8. Indus writing hieroglyphs in Mesopotamian artefacts

Functions of many seals to denote source of product and 'for approval' trade transactions

Archaeological framework for metals trade at Susa/Mesopotamia

Indus writing used hieroglyphs: context bronze-age

Trough as a hieroglyph

Archaeology and language: Archaeological context of Indus writing, ca. 3500 BCE

Indus writing corpora and evidence related to proto-Indian or proto-Indic or Indus language : Meluhha (mleccha)

Daha, dasyu, 'people'

Decoding Salut seal with Indian hieroglyphs (Indus script)

Annex A: Indus writing hieroglyphs

    The artisans' guild from Indus (Meluhha) assumed the form of a multi-national corporation, attested by *harosheth hagoyim*, **[cognate:** *kharoṣṭī goy* (Meluhha/mleccha)] 'smithy of nations' mentioned in the Old Testament. It appears that the Meluhhans were in contact with many interaction areas, Dilmun and Susa (elam) in particular. There is evidence for Meluhhan settlements outside of Meluhha. It is a reasonable inference that the Meluhhans with bronze-age expertise of creating arsenical and bronze alloys and working with other metals constituted the 'smithy of nations', Harosheth Hagoyim.

This hypothesis is confirmed by harosheth, (cognate *kharoṣṭī*) tradition. *kharoṣṭī* was a syllabic writing system with intimations of contacts with Aramaic writing system. Though early evidences of *kharoṣṭī* documents are dated to ca. 3$^{rd}$ century BCE, it is likely that some form of contract documentation using a proto-form of *kharoṣṭī* was perhaps used by artisan and traders, across a vast interaction area which covered a wide geographic area from Kyrgystan (Tocharian) to Haifa (Israel, Seaport on Mediterranean Ocean) – across Sarasvati-Sindu river-basins, Tigris-Euphrates doab, Caspian Sea, and Mediteranean Ocean – of three civilizations Indus, Mesopotamia and Egypt. The evidence of about 6000 Indus script inscriptions provides the details of products traded in this *harosheth hagoyim*, a smithy of nations, indeed. Harosheth is spelt in pronunciation: *khar-o'-sheth*. *Harosheth* and cognate *kharoṣṭī* may mean 'workmanship' or 'art of writing', apart from connoting specifically blacksmiths' writing system. Artisans had invented early writing systems necessitated by the economic imperative of bronze-age trade. In this smithy of nations, language was not a barrier. The barrier had been bridged by the invention and use of hieroglyphic and syllabic writing systems to record guild production and sea-faring or land-caravan trade transactions.

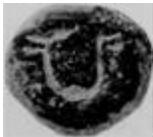 Seal. Daimabad1 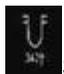 Sign342

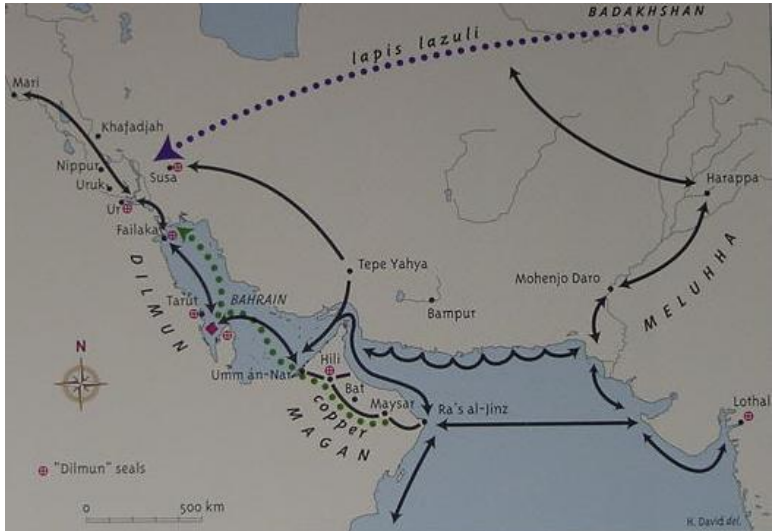

*kaṇḍ kanka* 'rim of jar' (Santali); rebus: *kaṇḍ* 'stone (ore) metal'. *karṇaka* 'rim'; rebus: 'scribe, account'. kárṇaka m. ' projection on the side of a vessel, handle ' ŚBr. [kárṇa -- ] Pa. kaṇṇaka -- ' having ears or corners '; Wg. Kaṇə ' ear — ring ' NTS xvii 266; S. kano m. ' rim, border '; P. kannā m. ' obtuse angle of a kite ' (→ H. kannā m. ' edge, rim, handle '); N. kānu ' end of a rope for supporting a burden '; B. kāṇā ' brim of a cup ', G. kānɔ m.; M. kānā m. ' touch — hole of a gun '.(CDIAL 2831). Rebus: கணக்கு kaṇakku , n. cf. gaṇaka. [M. kaṇakku.]

1. Number, account, reckoning, calculation, computation. This hieroglyph announces the arrival of a new professional, an expert carver who can keep accounts of the industrial goods produced in guild workshops and sorted out or displayed on circular working platforms. This hieroglyph is often the terminal signature tune of many inscriptions conveying the message that goods tallied using tablets have been consolidated together to create seal impressions as bills of lading for multi-commodity trade loads. The invention of writing has created a new professional: (accountant) scribe.

Contacts between users of Aramaic- *kharoṣṭī* writing systems may be seen as a continuum of interactions among Mesopotamian settlements and Meluhhan settlements as broadly indicated in the following map:

After Walter Reinhold Warttig Matted y de la Torre, 2005, Sumerian Dilmun
http://www.bibleorigins.net/dilmunmapseriduurseashorepersiangulf.html

**Evolving bronze-age and use of tallies, tokens, bullae for archives**

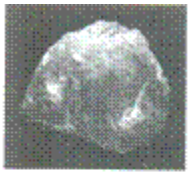

The lapidary had graduated into a smithy worker in the bronze age and needed Indus script inscriptions to account for processing, collating, and dispatching trade loads to trade contacts in interaction areas. The scribe created a seal to account for contributions by artisans of the guild as a record of product descriptions sorted, grouped and delivered into the treasury.

**Use of tallies, tokens for archives in the Near East**

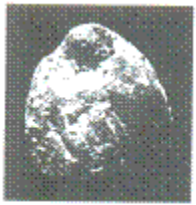 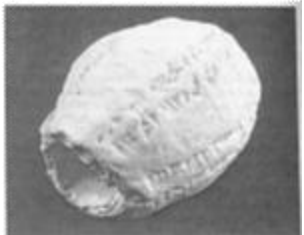

*"From the beginnings in about 30,000 BCE, the evolution of information processing in the prehistoric Near East proceeded in three major phases, each dealing with data of increasing specificity. First, during the Middle and late Upper Paleolithic, ca. 30,000 – 12,000 BCE, tallies referred to one unit of an unspecified item. Second, in the early Neolithic, ca. 8000 BCE, the tokens indicated a precise unit of a particular good. With the invention of writing, which took place in the urban period, ca. 3100 BCE, it was possible to record and communicate the name of the sponsor/recipient of the merchandise, formerly indicated by seals…The events that followed the invention of tokens can be reconstructed as follows: ca. 3700-2000 BCE: A second stage was reached when groups of tokens*

*representing particular transactions were enclosed in envelopes to be kept in archives. Some envelopes bore on the outside the impression of the tokens held inside. Such markings on envelopes were the turning point between tokens and writing. Ca. 3500-3100 BCE (starting in Uruk VI-V): Tablets displaying impressed markings in the shape of tokens superseded the envelopes. Ca. 3100-3000 BCE (starting in Uruk Iva): Pictographic script traced with a stylus on clay tablets marked the true takeoff of writing. The tokens dwindled…The tokens were mundane counters dealing with food and other basic commodities of life, but they played a major role in the societies that adopted them. They were used to manage goods and they affected the economy; they were an instrument of power and they created new social patterns; they were employed for data manipulation and they changed a mode of thought. Above all, the tokens were a counting and record-keeping device and were the watershed of mathematics and communication."*

(Denise Schmandt-Besserat, 1996, *How writing came about*, University of Texas Press, p.99, p. 125).

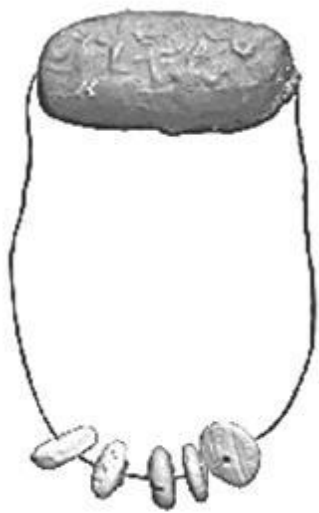

Tallies and tokens were used to archive records of counted goods which were basic necessities such as: animals (lamb, sheep, ewe, cow, dog); foods (bread, oil, food, sweet (honey?), beer, sheep's milk); textiles (textile, wool, type of garment or cloth, fleece, rope, type of mat or rug); commodities (perfume, metal, bracelet, ring, bed); service (make, build).

**Decoding of the identical inscription on the three tablets of Kanmer**

Tokens strung together with a bulla to constitute an archive. (After Denise Schmandt-Besserat, 1996, *How writing came about*, University of Texas Press).

Duplicate seal impressions are one type of tablets. An evidence for the use of such tablets as category tallies of lapidary workshops is provided by the finds at Kanmer. (Source: http://www.antiquity.ac.uk/projgall/agrawal323/Antiquity, D.P. Agrawal et al, Redefining the Harappan hinterland, *Anquity*, Vol. 84, Issue 323, March 2010)

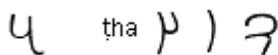

Indus      2   1   20

Obverse of these tiny 2 cm. dia. tablets show some incised markings. It is unclear from the markings if they can be compared with any glyphs of the script corpora. They may be 'personal' markings like 'potter's marks' – designating a particular artisan's workshop (working platform) or considering the short numerical strokes used, the glyphs may be counters (numbers or liquid or weight measures). More precise determination may be made if more evidences of such glyphs are discovered. Excavators surmise that the three tablets with different motifs on the obverse of the three tablets suggest different users/uses. They may be from different workshops of the same guild but as the other side of the tables showed, the product taken from three workshops is the same.

It is possible that the markings on the obverse of the three Kanmer tablets (as tallies) were markings using a form of *kharoṣṭī* proto-syllabary as follows, possibly indicting some quantitative measures of the products delivered to the furnace account scribe of turned (forged) native metal : *kharoṣṭī* numeral 'twenty' *kharoṣṭī* numeral 'two'. *kharoṣṭī* numeral 'one'. *kharoṣṭī* syllable (*ṭha-* for *ṭhakkura* 'blacksmith'?)

Glyph: One long linear stroke. koḍa 'one' (Santali) Rebus: koḍ 'artisan's workshop' (Kuwi) Glyph: meḍ 'body' (Mu.) Rebus: meḍ 'iron' (Ho.) Ligatured glyph : aḍar 'harrow' Rebus: aduru 'native metal' (Kannada). Thus the glyphs can be read rebus. Glyph: koḍiyum 'heifer' (G.) Rebus: koḍ 'workshop (Kuwi)

Glyph: sangaḍa 'lathe' (Marathi) Rebus 1: Rebus 2: sangaḍa 'association' (guild). Rebus 2: sangatarāsu 'stone cutter' (Telugu). The output of the lapidaries is thus described by the three tablets: *aduru meḍ sangaḍa koḍ* 'iron, native metal guild workshop'.

**Conjecturing a parallel with Sumer bulla envelope system**

The three perforated tablets (seal impressions) of Kanmer might have been strung together and the account compiled by the guild scribe to prepare a bill of lading. It is also possible that a seal impression on a bulla might have authenticated the bill of lading together with the three tablets (seal impressions) of Kanmer.

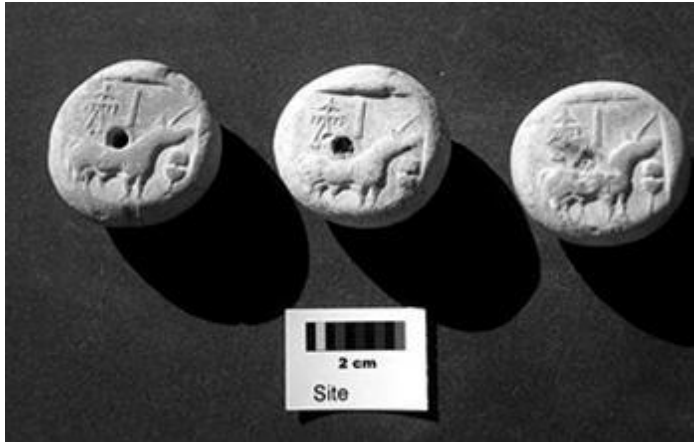

The hole on the following tablets may also have been strung together to create a tally of products delivered into the warehouse 'for approval', जांगड [jāṅgaḍa] (Marathi). [Note: The Kanmer archaeological report is scheduled for release at Udaipur on 18 April 2012, the World Heritage Day (Private Communication from Jeewan Kharakwal).]

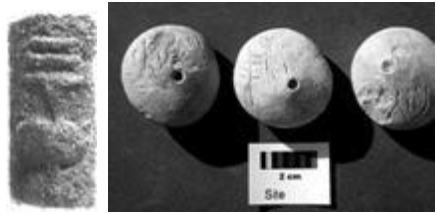

The practice of combining *kharoṣṭī* syllabary for names together with Indian hieroglyphs (from Indus script) for substantive messaging of the mint repertoire continues in the historical periods as evidenced by thousands of punch-marked coins starting from c. 600 BCE.

Glyph of standard device in front of the one-horned heifer: sãgāḍī lathe (Tu.)(CDIAL 12859). sāṅgaḍa That member of a turner's apparatus by which the piece to be turned is confined and steadied. सांगडीस धरणें To take into linkedness or close connection with, lit. fig. (Marathi) सांगाडी [ sāṅgāḍī ] f The machine within which a turner confines and steadies the piece he has to turn. (Marathi)सगडी [ sagaḍī ] f (Commonly शेगडी) A pan of live coals or embers. (Marathi) san:ghāḍo, saghaḍī (G.) = firepan; saghaḍī, śaghaḍi = a pot for holding fire (G.)[*culā sagaḍī* portable hearth (G.)]
h739B

Thus, the entire set of glyphs on the h1682A seal [denoting the heifer + standard device] can be decoded: koḍiyum 'heifer'; [kōḍiya] kōḍe, kōḍiya. [Tel.] n. A bullcalf. . k* దూడA young bull. Plumpness, prime. తరుణము. జోడుకోడయలు a pair of bullocks. Kōḍe adj. Young. Kōḍe-kāḍu. n. A young man.పడుచువాడు. [kārukōḍe] kāru-kōḍe. [Tel.] n. A bull in its prime. खोंड [khōṇḍa] m A young bull, a bullcalf. (Marathi) గోద [ gōda ] gōda. [Tel.] n. An ox. A beast. Kine, cattle.(Telugu) koḍiyum (G.) rebus: koḍ 'workshop' (G.) B. kōdā 'to turn in a lathe'; Or. kŭnda 'lathe', kūdibā, kū̃d 'to turn' (→ Drav. Kur. kū̃d 'lathe') (CDIAL 3295)

**Conclusion**

Tokens designed to count goods evolved over millennia into hieroglyphs to represent words denoting the bronze-age goods and processes. This stage of rebus representation of sounds of words of meluhha (mleccha language) was the stage penultimate to the culminating stage which used representation of syllables graphically in Brahmi and Kharoshti scripts. This culmination of the process for literacy and civilization was the contribution made by artisans of the bronze-age of Sarasvati civilization (also called Indus civilization).

**S. Kalyanaraman**

Sarasvati Research Center

Herndon, VA 20171

April 17, 2012 kalyan97@gmail.com

(Gherardo Gnoli) Originally Published: December 15, 1993

---

[1] Kalyanaraman, S. 2012. *Indian Hieroglyphs – Invention of Writing.* Herndon: Sarasvati Research Center
[2] Schmandt-Besserat, D. 1996. *How Writing Came About.* Austin: The University of Texas Press.
[3] Schmandt-Besserat, D. (1992), *Before Writing*, 2 vols. Austin: The University of Texas Press.

[4] Ibid.

[5] Schmandt-Besserat, D. 2009. SCRIPTA, Volume 6 (September 2009): 145
[6] Kalyanaraman, S. 2012. *Indian Hieroglyphs – Invention of Writing.* Herndon: Sarasvati Research Center

[7] Nissen, H.J., Damerow, P., Englund, R.K., 1993. *Archaic Bookkeeping*, Chicago: The University of Chicago Press, pp. 64-65.